\documentclass{triton}

\usepackage{amsmath,amsfonts,bm}

\def\eqref#1{equation~\ref{#1}}

\def\1{\bm{1}}

\DeclareMathAlphabet{\mathsfit}{\encodingdefault}{\sfdefault}{m}{sl}
\SetMathAlphabet{\mathsfit}{bold}{\encodingdefault}{\sfdefault}{bx}{n}

\usepackage{amsmath,amsfonts,amssymb,mathtools,bm}  
\usepackage{thmtools}          
\usepackage{algorithm}         
\usepackage{algpseudocode}     
\usepackage{enumitem}          
\usepackage{siunitx}           
\usepackage{tikz}              
\usepackage{booktabs}          
\usepackage{subcaption}        

\usepackage[utf8]{inputenc} 
\usepackage[T1]{fontenc}    
\usepackage{hyperref}       
\usepackage{url}            

\usepackage{xurl}          
\usepackage{booktabs}       
\usepackage{amsfonts}       
\usepackage{nicefrac}       
\usepackage{microtype}      
\usepackage{graphicx}
\usepackage{subcaption} 
\usepackage{xspace}
\usepackage[frozencache,cachedir=.]{minted}
\usepackage{longtable}

\usepackage{nicematrix}
\usepackage{multirow}
\usepackage{makecell}
\usepackage{soul}  
\usepackage{wrapfig}
\usepackage{enumitem}

\usepackage{colortbl}

\definecolor{maroon}{HTML}{F26035}
\definecolor{yellow}{HTML}{FDBC42}
\definecolor{lavender}{HTML}{734f96}
\definecolor{ai2pink}{HTML}{ED017D}%
\definecolor{ai2lightpink}{HTML}{FDDEEC}
\definecolor{ai2offwhite}{HTML}{faf2e9}

\definecolor{ai2green}{HTML}{0FCB8C}
\definecolor{ai2purple}{HTML}{B932EB}
\definecolor{ai2darkgreen}{HTML}{105257}

\newcommand{\ours}{\textsc{Guru}\xspace}

\definecolor{cellHighlight}{HTML}{dbefff}
\newcommand{\HL}{\cellcolor{cellHighlight}}

\newcommand{\hf}{\raisebox{-0.25em}{\includegraphics[height=1em]{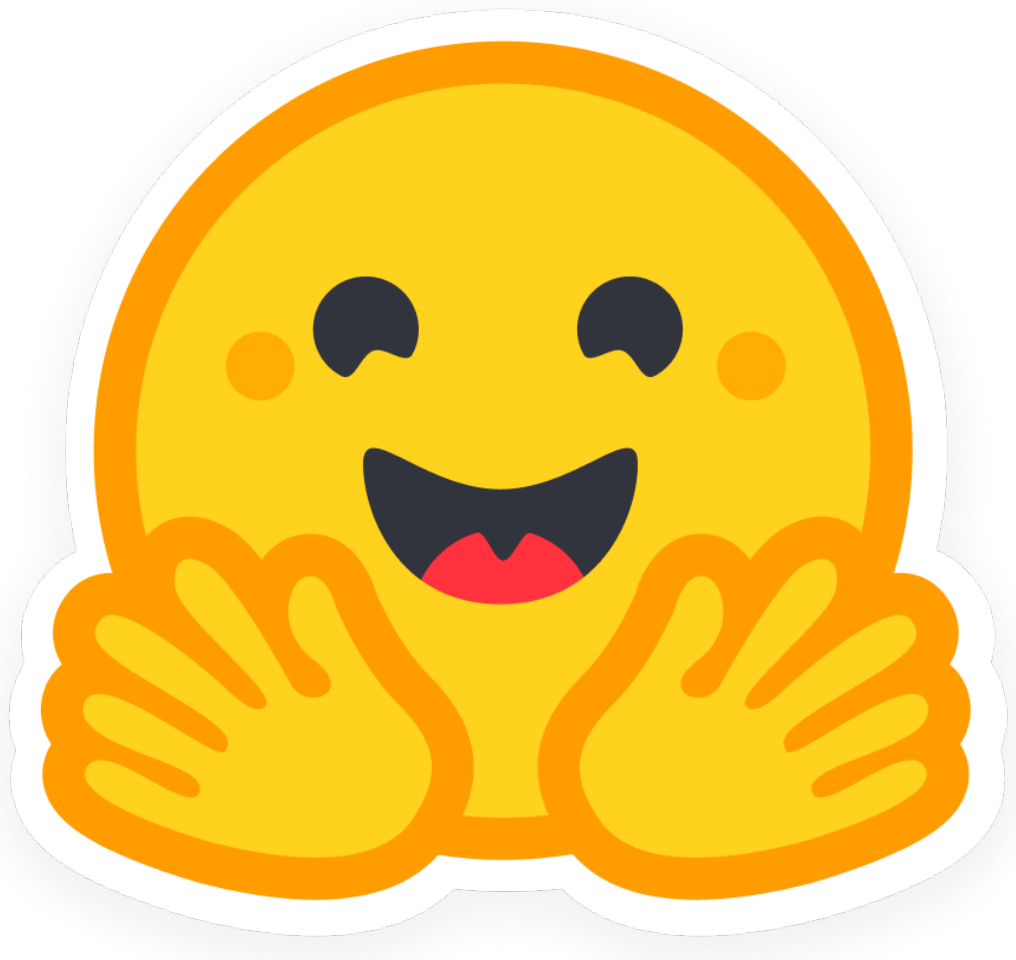}}}
\newcommand{\gh}{\raisebox{-0.25em}{\includegraphics[height=1em]{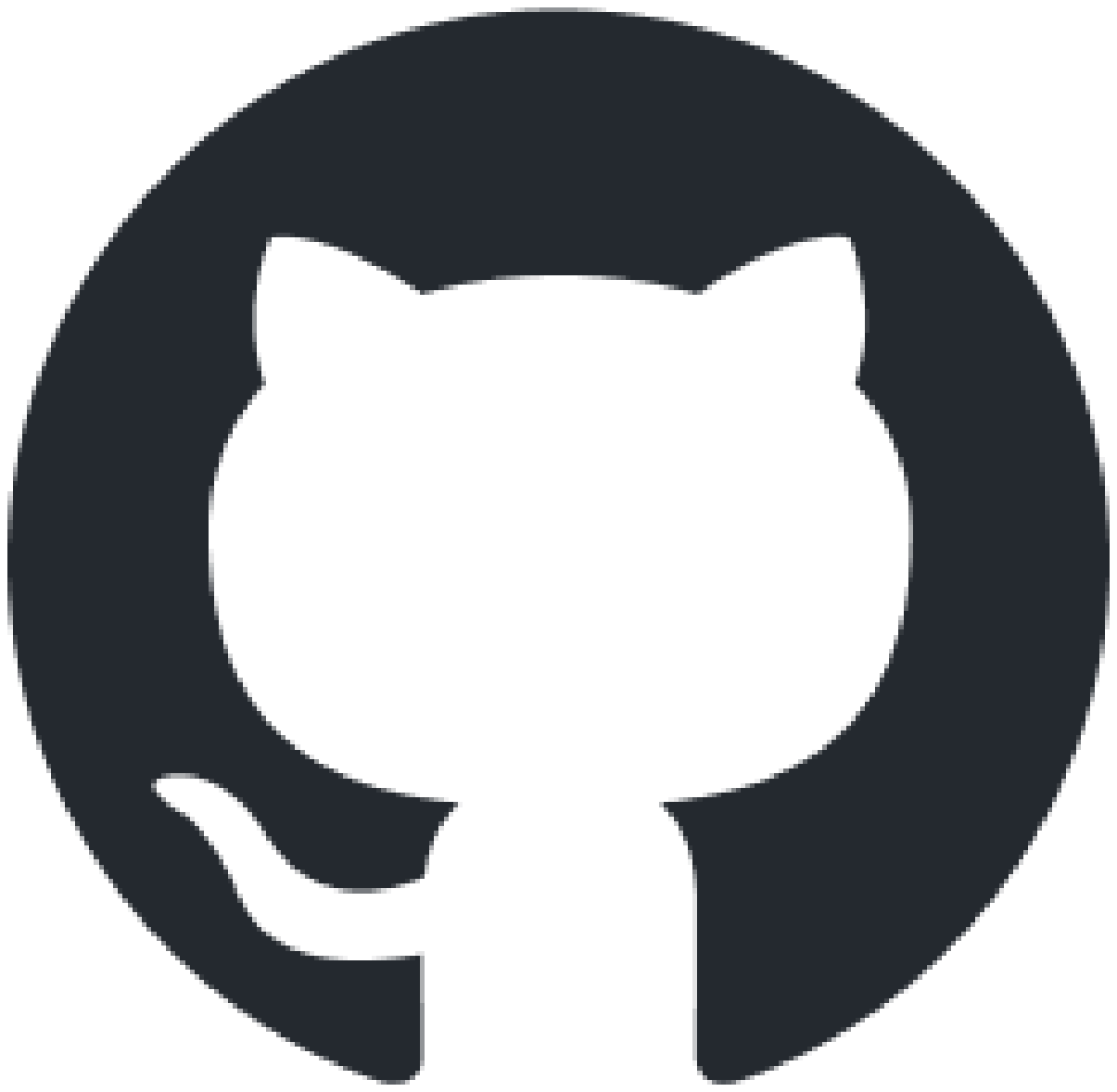}}}

\title{Revisiting Reinforcement Learning for LLM Reasoning \\ from A Cross-Domain Perspective}

\author[1,*]{Zhoujun Cheng}\authorsep{,}
\author[1,*]{Shibo Hao}\authorsep{,}
\author[1,*]{Tianyang Liu}\authorsep{\\} 
\author[2]{Fan Zhou}\authorsep{,}
\author[1]{Yutao Xie}\authorsep{,}
\author[1]{Feng Yao}\authorsep{,}
\author[1]{Yuexin Bian}\authorsep{,}
\author[3]{Yonghao Zhuang}\authorsep{,}
\author[4]{Nilabjo Dey}\authorsep{\\}
\author[1]{Yuheng Zha}\authorsep{,}
\author[1]{Yi Gu}\authorsep{,}
\author[1]{Kun Zhou}\authorsep{,}
\author[2]{Yuqi Wang}\authorsep{,}
\author[3]{Yuan Li}\authorsep{,}
\author[2]{Richard Fan}\authorsep{,}
\author[2]{Jianshu She}\authorsep{\\}
\author[2]{Chengqian Gao}\authorsep{,}
\author[4]{Abulhair Saparov}\authorsep{,}
\author[2]{Haonan Li}\authorsep{,}
\author[2]{Taylor W. Killian}\authorsep{,}
\author[2]{Mikhail Yurochkin}\authorsep{\\}
\author[2]{Zhengzhong Liu}\authorsep{,}
\author[2,3]{Eric P. Xing}\authorsep{,}
\author[1]{Zhiting Hu}

\affiliation[1]{UC San Diego}\affiliationsep{,}
\affiliation[2]{MBZUAI}\affiliationsep{,}
\affiliation[3]{Carnegie Mellon University}\affiliationsep{,}
\affiliation[4]{Purdue University}

\contribution[*]{Equal Contribution}

\abstract{

Reinforcement learning (RL) has emerged as a promising approach to improve large language model (LLM) reasoning, yet most open efforts focus narrowly on math and code, limiting our understanding of its broader applicability to general reasoning. 
A key challenge lies in the lack of reliable, scalable RL reward signals across diverse reasoning domains.
We introduce \ours, a curated RL reasoning corpus of 92K verifiable examples spanning six reasoning domains—Math, Code, Science, Logic, Simulation, and Tabular—each built through domain-specific reward design, deduplication, and filtering to ensure reliability and effectiveness for RL training.
Based on \ours, we systematically revisit established findings in RL for LLM reasoning and observe significant variation across domains. For example, while prior work suggests that RL primarily elicits existing knowledge from pretrained models, our results reveal a more nuanced pattern: domains frequently seen during pretraining (Math, Code, Science) easily benefit from cross-domain RL training, while domains with limited pretraining exposure (Logic, Simulation, Tabular) require in-domain training to achieve meaningful performance gains, suggesting that RL is likely to facilitate genuine skill acquisition.
Finally, we present \ours-7B/32B, two models that achieve state-of-the-art performance among open models RL-trained with publicly available data, outperforming best baselines by 7.9\% and 6.7\% on our 17-task evaluation suite across six reasoning domains. We also show that our models effectively improve the Pass@\emph{k} performance of their base models, particularly on complex tasks less likely to appear in pretraining data. We release data, models, training and evaluation code to facilitate general-purpose reasoning research at \small{\url{https://github.com/LLM360/Reasoning360}}.
}

\begin{document}
\maketitle


\begin{figure*}[!htp]
\centering
\begin{subfigure}[t]{0.45\textwidth}
        \centering
        \includegraphics[width=\linewidth]{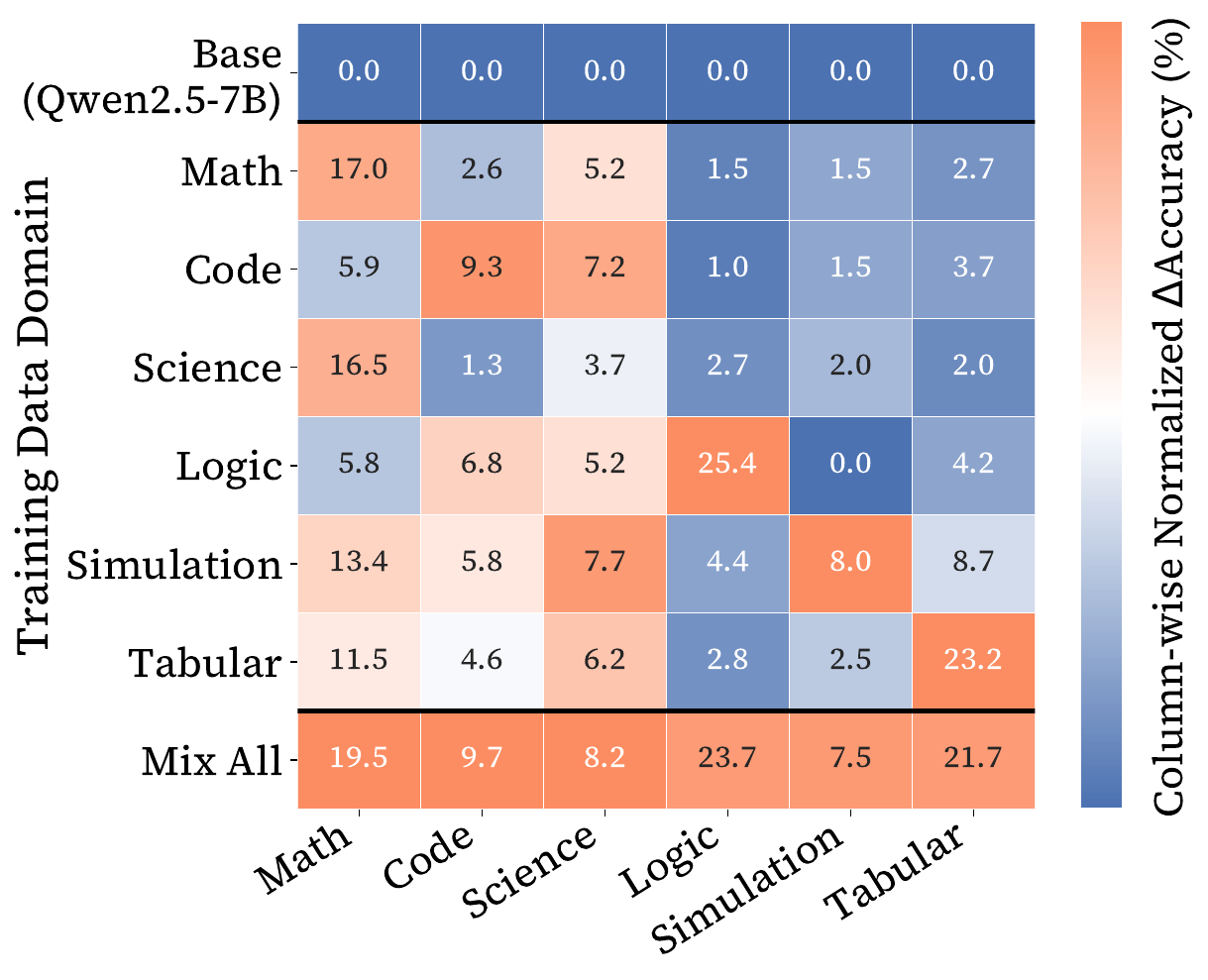}
        \caption{Cross-domain transfer performance.}
        \label{fig:sub-a}
    \end{subfigure}
    \hfill
    \begin{subfigure}[t]{0.49\textwidth}
        \centering
        \includegraphics[width=\linewidth]{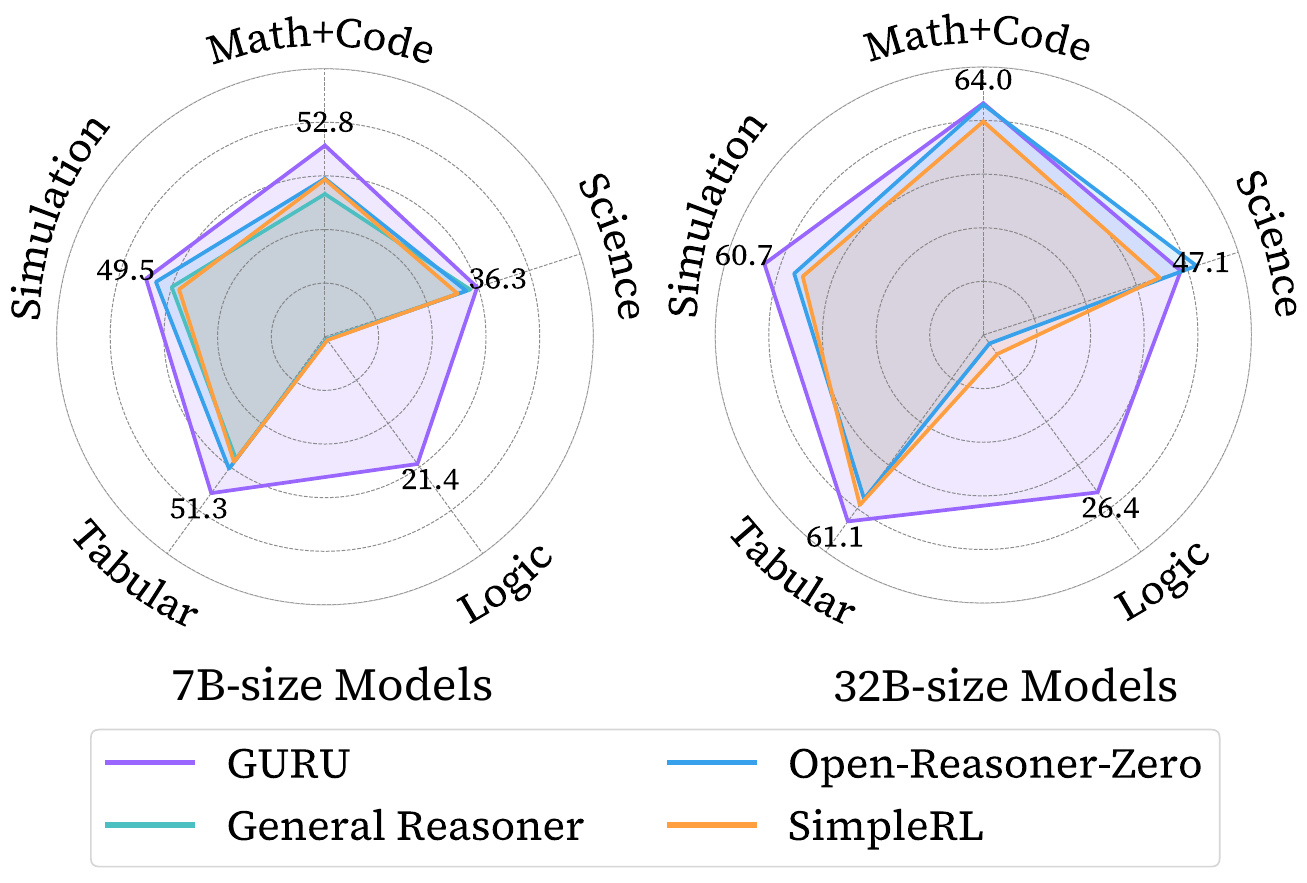}
        \caption{Model performance across reasoning domains.}
        \label{fig:sub-b}
    \end{subfigure}
    \caption{
    \textbf{Left:} Absolute accuracy gain (\%) over the base model when RL-trained on various domains. Pretrained-heavy domains (Math, Code, Science) benefit from cross-domain training, while others require in-domain data, indicating RL aids skill acquisition.
    \textbf{Right:} Our \ours-7B/32B models consistently outperform strong open baselines across 17 reasoning tasks when RL-trained with mixed-domain data. 
    } 

\label{fig:intro}
\end{figure*}

\newpage
\section{Introduction}\label{sec:intro}

Recent frontier reasoning models, such as OpenAI-O3~\citep{openaio3} and DeepSeek-R1~\citep{guo2025deepseekr1}, demonstrate impressive performance and generalizability \textit{across diverse domains}.
These advances heavily leverage reinforcement learning (RL) as a core post-training technique to enhance large language model~(LLM) reasoning capabilities.
Motivated by these results, the open-source community has developed competitive reasoning models using RL~\citep{yu2025dapo,hu2025open,wang2025octothinker} and systematically studied the principles behind RL's effectiveness for reasoning~\citep{zeng2025simplerl,yue2025does,shao2025spurious,min-entropy, one-shot-rl}.

However, the vast majority of RL works for LLM reasoning train and evaluate RL models \textit{exclusively} on math and code domains. The domain-specific concentration raises \textit{two critical limitations} in understanding and improving general-purpose reasoning. First, the current understanding of RL's role in reasoning is predominantly derived from studies focused on math and coding~\citep{spuriousreward,zeng2025simplerl,yue2025does,one-shot-rl,echo-chamber}, potentially limiting generalizability. For instance, recent observations about RL's fundamental mechanisms—whether it teaches new reasoning skills or merely elicits existing knowledge—\textit{may not generalize} across other domains.
Second, this approach produces models that excel \textit{only within their specialized training domains}~\citep{hu2025open,zeng2025simplerl,yu2025dapo,deepscaler2025}, limiting both our understanding of how RL enables reasoning and practical performance gains across diverse reasoning tasks. Consequently,  studying RL across multiple domains is essential for advancing general reasoning research and deepening our understanding of RL’s contributions.

A key bottleneck in extending RL-based reasoning to diverse domains is the scarcity of reliable reward signals and curated training data. Math and code problems are more widely sourced and studied, and offer easily verifiable answers~\citep{li2024numinamath,likaixin2024taco-verified}. However, reasoning domains beyond them usually require more careful and creative data sourcing and domain-specific reward design. Even for math and coding domains, where queries and reward signals are more readily available, publicly available datasets require extensive curation to address common issues: duplicates, noisy samples, and unbalanced difficulty distributions~\citep{llama-nemotron,orz}.

We introduce \ours, a curated RL corpus spanning six diverse reasoning domains: Math, Code, Science, Logic, Simulation, and Tabular. Each domain within \ours is constructed via a meticulous data pipeline designed to ensure reliability and utility for RL. The pipeline involves: data sourcing and synthesis, deduplication, domain-specific reward function design (with rule, execution, or model-based verification) for precise feedback, and both heuristic and model-based filtering to remove noisy or trivial examples and to ensure an appropriate difficulty distribution. The final corpus offers 92k examples, each paired with a verifiable reward signal, establishing a foundational and openly available benchmark to systematically conduct research on general-purpose reasoning with RL.

By performing RL on six domains in \ours with \texttt{Qwen2.5-7B} and \texttt{32B} base models, we revisit established insights in RL-based reasoning and observe that these conclusions are \textit{domain-dependent}.
Most critically, prior work suggests RL primarily elicits existing knowledge from pretrained models rather than learning new capabilities, even with indirect or minimal elicitation like wrong or random labels or even performing RL from a single example~\citep{echo-chamber,yue2025does,spuriousreward,one-shot-rl}. As shown in Figure~\ref{fig:intro}(Left), domains frequently seen during continual pretraining~\citep{ke2023continual,yang2024qwen2} (Math, Code, Science) readily show performance gains from cross-domain RL training (\textcolor{orange}{warm colors} on non-diagonal cells), indicating reasoning capabilities are more easily elicited in these domains even with cross-domain data. In contrast, relatively less familiar domains (Logic, Simulation, Tabular) require in-domain RL training to achieve meaningful improvements (\textcolor[rgb]{0.2,0.4,0.7}{cold colors} on all non-diagonal cells) suggesting RL is more likely to learn new skills in these cases~(\Cref{sec:cross_domain_transfer}).
Additionally, we find that response length increases correlate with performance gains only in certain domains (\Cref{subsec:rrld}), and difficulty filtering improves within-domain performance while harming cross-domain transfer (\Cref{subsec:etdd}).
These findings reveal that RL reasoning mechanisms are more domain-specific than previously recognized, emphasizing the need for multi-domain training and evaluation in RL for LLM reasoning research.

Finally, we deliver \ours-7B and \ours-32B, two general reasoning models trained with RL on the \ours dataset. On our unified evaluation suite with 17 reasoning tasks across six domains, our models establish a new state-of-the-art among open models trained with publicly available data, outperforming the best baselines by 7.9\% and 6.7\% respectively.
Furthermore, we observe that Pass@\emph{k} patterns of RL-trained and base models, which is defined as the LLM's reasoning boundary in  \citet{yue2025does}, vary by tasks, model scales, and generation parameters (e.g., temperature and top-p) (\Cref{subsec:passk_analysis}). For example, on a synthetic Zebra Puzzle task, a complex constraint satisfaction problem that is less likely to appear in pretraining data, both \ours-7B and \ours-32B effectively expand the reasoning boundary of their base models.
Overall, these substantial performance gains and cross-domain findings demonstrate the importance of multi-domain RL research for advancing general reasoning capabilities. We release the \ours dataset, models, and code at {\small{\url{https://github.com/LLM360/Reasoning360}}} for the community's use and continued study.

\section{Data Construction}\label{sec:dataconstruction}

\begin{wrapfigure}{r}{0.35\textwidth}
    \centering
    \includegraphics[width=0.95\linewidth]{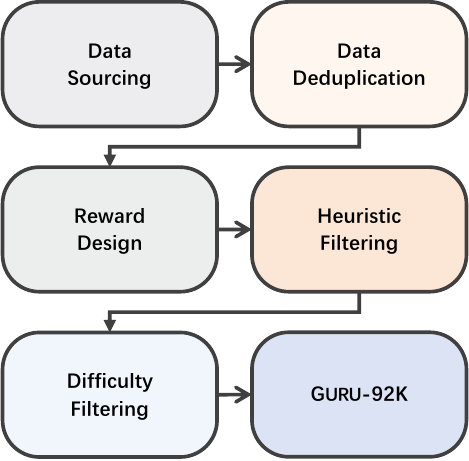}
    \caption{Overview of the data curation pipeline of \ours dataset.}
    \label{fig:pipeline}
\end{wrapfigure}

Datasets used for RL reasoning predominantly focus on narrow domains such as Math~\citep{deepscaler2025, hu2025open, skywork-or1-2025, zeng2025simplerl} and Code~\citep{code-r1, deepcoder2025}. However, applying RL on such single-domain datasets often leads to overfitting to domain-specific structures and heuristics. As illustrated in~\Cref{fig:intro}, models trained to excel on advanced math benchmarks suffer sharp performance drops when evaluated on other domains. Even datasets that claim to span general reasoning capabilities tend to remain confined within the STEM boundary and exhibit similar generalization failures. Furthermore, existing RL reasoning datasets frequently contain redundant samples, noisy queries, and poorly calibrated difficulty levels. These limitations highlight the need for principled data curation, including deduplication, domain-specific filtering, and difficulty-aware sampling. To address these challenges, we propose a comprehensive data curation pipeline for constructing \ours{}, a multi-domain RL dataset designed to promote both domain diversity and reward verifiability. An overview of our pipeline is shown in~\Cref{fig:pipeline}, consisting of five stages detailed below.

\subsection{Data Sourcing} 
We curate data across six reasoning-intensive domains, ensuring both diversity and verifiability:

\begin{itemize}[leftmargin=1.25em,itemindent=0.15em,labelsep=0.2em,itemsep=0.05em]
    \item \textbf{Math:} We adopt recent math reasoning collections including \emph{OR1}~\citep{skywork-or1-2025}, \emph{DAPO}~\citep{yu2025dapo}, and \emph{DeepScaler}~\citep{deepscaler2025}, which compile numerous math datasets, notably previous competition problems such as AIME and AMC.
    
    \item \textbf{Code:} We include programming problems sourced from online coding platforms, programming competitions, and synthetic code generation tasks. Our collection features real-world problems from \emph{LeetCode}~\citep{leetcode}, curated and verified problems from \emph{TACO-Verified}~\citep{likaixin2024taco-verified}, synthetic tasks from \emph{PrimeIntellect}~\citep{2025synthetic1}, and historical problems from \emph{LiveCodeBench}~\citep{jain2024livecodebench}. Notably, we reuse filtered subsets of \emph{PrimeIntellect} and \emph{LiveCodeBench} as processed by \emph{DeepCoder}~\citep{deepcoder2025}.
    
    \item \textbf{Science:} We adopt \emph{WebInstruct-Verified}~\citep{generalreasoner}, a dataset sourced from web content and refined using LLMs, as the primary source for science questions.
    
    \item \textbf{Logic:} We include symbolic reasoning tasks sourced from existing datasets and newly synthesized tasks. From existing datasets, we include \emph{ARC-AGI} (1 and 2)~\citep{chollet2024arc}, which focus on abstract rule induction over symbolic grids, and \emph{BARC}~\citep{li2024combininginductiontransductionabstract}, which extends these tasks for inductive and transductive generalization. In addition, we synthesize three tasks for RL training: \emph{Zebra Puzzle}~\citep{logiczebra}, a classic logic grid puzzle requiring models to chain positional and equality constraints; \emph{Ordering Puzzles}, which require recovering a unique linear order of up to 50 objects from diverse relational constraints; and \emph{Graph Search}, where models must construct a predicate graph from natural language facts and reason over it to identify entity paths.
    
    \item \textbf{Simulation:} We include simulation-style reasoning tasks from \emph{Code I/O (PyEdu)}~\citep{li2025codei}, where models must predict outputs or infer inputs based on code snippets without execution.
    
    \item \textbf{Tabular:} We repurpose existing table-based QA datasets involving both single- and multi-table reasoning. Specifically, we adopt \emph{HiTab}~\citep{cheng2021hitab} and \emph{MultiHierTT}~\citep{zhao2022multihiertt}.
\end{itemize}

\Cref{tab:sourcing} summarizes the concrete data statistics each domain. Further details, including specific data links, collection or synthesis methods, and curation procedures for all datasets, are provided in \Cref{sec:data_appendix}.

\subsection{Data Deduplication}
Significant content overlap exists within the Math and Code domains, as many datasets originate from shared upstream sources. For example, numerous Math datasets reuse problems from past AIME exams or curated collections such as \emph{NuminaMath}~\citep{li2024numinamath}. Likewise, Code datasets are often derived from online coding platforms (e.g., \emph{LeetCode}) and curated benchmarks (e.g., \emph{APPS}, \emph{CodeContests}), leading to repeated or near-duplicate problems across datasets. In our preliminary analysis, we observe that similarity-based metrics—such as embedding distance, Jaccard similarity, and n-gram overlap—are prone to false positives. For instance, n-gram matching may erroneously flag distinct samples that merely share a common prefix or structural template. To mitigate this, we apply a conservative deduplication strategy: if one question is a strict substring of another, the shorter sample is removed. This approach eliminates 27.2\% of Math samples and 7.5\% of Code samples. To further support transparency and data quality assessment, we construct metadata-level mappings of dataset provenance and cross-source dependencies. Visualizations of these relationships, including Sankey diagrams for Math and Code datasets, are provided in \Cref{sec:data_appendix}.

\begin{table}[!t]
\caption{\ours dataset statistics across all 6 domains, including raw sizes and filtered subsets.}
\centering
\label{tab:sourcing}
\resizebox{0.95\textwidth}{!}{
\setlength{\tabcolsep}{10pt}
\begin{NiceTabular}{l|lrcc}
\toprule
\textbf{Domain} & \textbf{Dataset} & \textbf{Raw Size} & \textbf{\makecell{w/ Dedup.\\\& Domain Filt.}} & \textbf{\makecell{w/ Difficulty\\Filt.}} \\
\midrule

\multirow{3}{*}{\textbf{Math}} 
  & OR1~\citep{skywork-or1-2025}           & 105k  & \multirow{3}{*}{\makecell{118.2k\\ {\tiny{(-27.2\%)}} }} & \multirow{3}{*}{\makecell{54.4k\\{\tiny{(-54.0\%)}}}} \\
  & DAPO~\citep{yu2025dapo}          & 17k   &                        & \\
  & DeepScaler~\citep{deepscaler2025}    & 40.3k &                        & \\
\midrule

\multirow{4}{*}{\textbf{Code}} 
  & LeetCode~\citep{leetcode}         & 2k     & \multirow{4}{*}{\makecell{23.7k\\{\tiny{(-26.2\%)}}}} & \multirow{4}{*}{\makecell{18.1k\\{\tiny{(-23.6\%)}}}} \\
  & TACO-Verified~\citep{likaixin2024taco-verified}    & 12.9k  &                        & \\
  & PrimeIntellect~\citep{2025synthetic1}   & 16.3k  &                        & \\
  & LiveCodeBench (history)~\citep{jain2024livecodebench}    & 0.9k   &                        & \\
\midrule

\textbf{Science} & WebInstruct-Verifed~\citep{generalreasoner} & 232k & \makecell{31.7k\\{\tiny{(-86.3\%)}}} & \makecell{3.6k\\{\tiny{(-88.6\%)}}} \\
\midrule

\multirow{6}{*}{\textbf{Logic}} 
  & Zebra Puzzle       & 5.7k   & \multirow{6}{*}{\makecell{13.6k\\{\tiny{(-12.8\%)}}}} & \multirow{6}{*}{\makecell{6.3k\\{\tiny{(-53.7\%)}}}} \\
  & Ordering Puzzle   & 2.9k   &                        & \\
  & Graph Puzzle      & 2.8k   &                        & \\
  & ARC-AGI~\citep{chollet2024arc}       & 0.4k   &                        & \\
  & ARC-AGI-2~\citep{chollet2025arcagi2newchallengefrontier}          & 1k   &                        & \\
  & BARC~\citep{li2024combininginductiontransductionabstract}             & 3.4k   &                        & \\
\midrule

\textbf{Simulation} & Code I/O (PyEdu)~\citep{li2025codei} & 227k & \makecell{12.1k\\{\tiny{(-94.7\%)}}} & \makecell{3.7k\\{\tiny{(-69.4\%)}}} \\
\midrule

\multirow{2}{*}{\textbf{Tabular}} 
  & HiTab~\citep{cheng2021hitab}            & 7.5k & \multirow{2}{*}{\makecell{10.3k\\{\tiny{(-36.7\%)}}}} & \multirow{2}{*}{\makecell{6.1k\\{\tiny{(-32.7\%)}}}} \\
  & MultiHierTT~\citep{zhao2022multihiertt}      & 7.8k &                       & \\
\midrule

\HL\textbf{Total} & \HL-- & \HL\textbf{684.9k} & \HL\textbf{209.6k} & \HL\textbf{91.9k} \\
\bottomrule
\end{NiceTabular}
}
\end{table}

\subsection{Reward Design} 
A key challenge in RL for reasoning tasks is designing reward functions that are fully automatic, low-noise, and domain-appropriate. Across all domains in our dataset, we adopt binary rewards: a sample receives a reward of 1 if the output is judged as correct under domain-specific verification rules, and 0 otherwise. We categorize reward design into three types, described below. Additional task-specific details about reward design are further provided in \Cref{sec:data_appendix}.

\begin{enumerate}
[leftmargin=1.25em,itemindent=0.15em,labelsep=0.2em,itemsep=0.05em]
    \item \textbf{Rule-Based Matching:} This is the most common strategy, employed in domains such as Math, Logic, Simulation, and Tabular reasoning. Despite surface-level differences, these tasks share a common structure: the model is prompted to produce its final answer in a designated format—typically enclosed in \texttt{\textbackslash boxed\{\}}, wrapped in a special tag (e.g., \texttt{<answer>}), or contained within a JSON block. The verifier extracts this region, normalizes the output, and applies strict matching. For Math tasks, we integrate a symbolic program~\citep{cui2025process} to handle algebraic and notational variations.
    
    \item \textbf{Execution-Based Verification:} This approach is used in Code domains, where correctness is determined by program behavior. The model generates a function or script, which is executed against a suite of test cases. A reward of 1 is given only if all test cases pass. For tasks using stdin/stdout formats, we incorporate fuzzy comparison routines to handle formatting inconsistencies and numerical tolerances.
    
    \item \textbf{Model-Based Verification:} This method is used for Science tasks, where answers are often open-ended and difficult to evaluate using deterministic rules. We employ a 1.5B verifier model~\citep{generalreasoner} to assess whether the model's response semantically entails the reference answer, enabling reward computation for tasks that require greater flexibility in interpretation.
\end{enumerate}

\subsection{Heuristic Filtering} 
Reliable reinforcement learning requires accurate and stable reward signals, which can be undermined by noisy, ambiguous, or poorly constructed samples. To address this, we implement a heuristic filtering stage across all domains, aimed at improving reward verifiability, stability, and coverage diversity. Concretely, we remove samples with overly long prompts that risk truncation, invalid reference outputs (e.g., code that fails its own test cases), or excessively large inputs that may hinder stable reward computation in multi-process environments. For structurally complex tasks (e.g., graph reasoning), we control instance difficulty to retain only those with sufficient symbolic or compositional depth. Additional domain-specific filtering heuristics, including thresholds for size, precision, and redundancy, are detailed in \Cref{sec:data_appendix}. 
After heuristic filtering, some categories may still exceed 10k examples. In such cases, we apply uniform random sampling to cap each subset at approximately 10k examples, ensuring a manageable dataset size and balanced task coverage.

\subsection{Difficulty Filtering}

To further enhance RL data quality and ensure that RL focuses on challenging and learnable samples, we introduce a \textit{difficulty-aware filtering stage}. This stage is designed to (i) prioritize examples that present sufficient reasoning difficulty, and (ii) aggressively remove samples that possibly exhibit annotation noise or unstable reward signals.

We implement this by measuring the empirical pass rates of a weak model ($M_{\text{weak}}$, \texttt{Qwen2.5-7B-Instruct}) and a strong model ($M_{\text{strong}}$, \texttt{Qwen3-30B-A8B}) over $N=16$ runs per sample, denoted as $P_{\text{weak}}$ and $P_{\text{strong}}$, respectively. Based on the statistics, we discard samples that satisfy any of the following conditions:

\begin{enumerate}
[leftmargin=1.25em,itemindent=0.15em,labelsep=0.2em,itemsep=0.05em]
    \item \textbf{Overly easy samples}: $P_{\text{weak}} \geq \frac{15}{16}$. Such examples are consistently solved by the weak model and thus provide limited signal for further improvement through reinforcement learning.
    
    \item \textbf{Potentially noisy samples}: $P_{\text{strong}} = 0$. These samples are consistently failed by the strong model, which often indicates that the task is either ambiguous, incorrectly labeled, or inherently unsuitable for RL training (e.g., questions requiring extreme numeric precision or external tools to solve accurately).
    
    \item \textbf{Anomalous samples}: $P_{\text{weak}} > P_{\text{strong}}$. These samples typically occur when the weak model consistently produces an answer that matches the provided label. Our empirical analysis suggests that such behavior often stems from incorrect ground-truth labels in the dataset. We suspect this arises because the weak model has previously encountered these flawed question–answer pairs during pretraining or supervised finetuning, leading it to memorize the incorrect response. These samples can mislead the RL process by reinforcing erroneous behaviors.
\end{enumerate}

To further refine the dataset in specific domains, we analyze the \textit{difficulty gap}, defined as $P_{\text{strong}} - P_{\text{weak}}$. This metric serves as a proxy for the \textit{learnability} of a sample, where higher values indicate greater potential for RL to improve model reasoning capabilities. In the Math domain, given its large scale and generally high baseline performance, we adopt a more aggressive filtering strategy. We remove samples where the difficulty gap is marginal ($P_{\text{strong}} - P_{\text{weak}} \leq \frac{6}{16}$) and the strong model already demonstrates high competence ($P_{\text{strong}} \geq 0.75$). This ensures that RL focuses on sufficiently challenging math problems where meaningful improvements remain achievable. In the Science domain, where data quality and model-based verifier are inherently noisier, we apply a stricter criterion by discarding samples with $P_{\text{strong}} - P_{\text{weak}} < 0.5$, prioritizing samples with clearer reasoning gaps and minimizing the inclusion of ambiguous or low-signal examples.


\section{Analysis of Cross-Domain RL for Reasoning}
\label{sec:analysis}

To gain a deeper understanding of how reasoning capabilities generalize across different domains with RL, we conducted a controlled experimental analysis with our \ours dataset. Specifically, we investigated the impact of RL on single reasoning domains versus a mixed-domain corpus on their subsequent performance across a suite of domain-specific benchmarks.  These experiments aim to inform the design of additional large-scale RL reasoning experiments while also providing a broadened perspective on the utility of data diversity when developing reasoning models. 

\begin{figure}[!t]
\centering
\includegraphics[width=\textwidth]{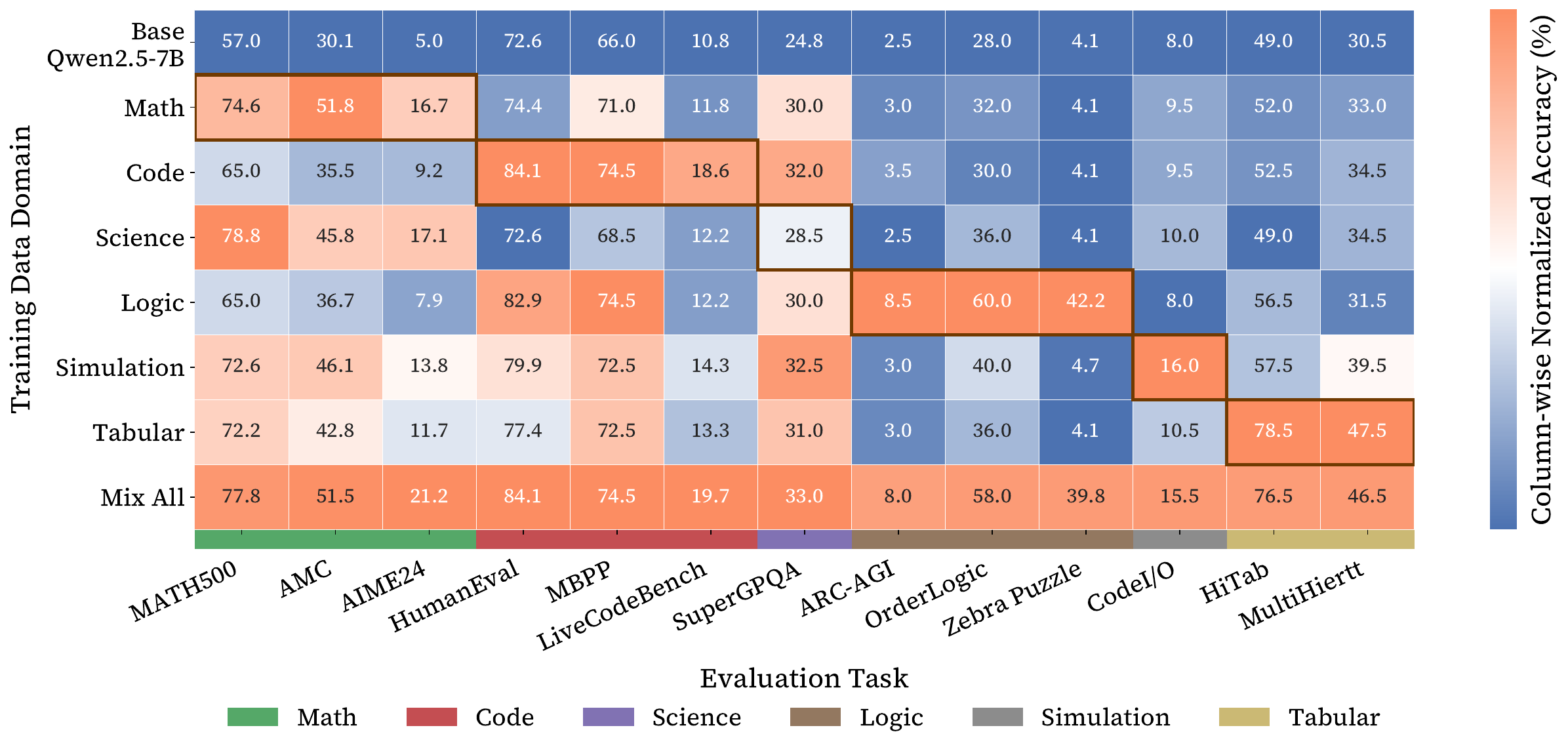}
\caption{\textbf{Cross-Domain RL Transfer Performance per Task.} 
    The heatmap shows the performance gains (accuracy) from RL training on different domains~(rows) when evaluated on the test sets on different domains~(columns). \textcolor{orange}{Warmer} colors indicate higher performance gains, computed by applying min-max normalization to the validation accuracies within each column. Accuracy is reported using the checkpoint with the highest average score across tasks.
    \textcolor{brown}{Khaki-colored} rectangles mark in-domain evaluations (diagonal); others reflect cross-domain generalization.
    This highlights differential transferability: Math, Code, and Science benefit significantly from cross-domain transfer, while Logic, Simulation, and Tabular tasks see limited gains, with improvements primarily driven by within-domain training. Notably, naive mixed-domain by combining data from all domains performs on par or better than single-domain RL.
}
\label{fig:cross_domain_heatmap}
\end{figure}

To this end, we constructed an experimental dataset by randomly sampling a subset of 3K training samples from each of the six domains covered by our \ours dataset. These subsets form a mixed training dataset totaling 18K samples called \ours-18K. We train reasoning models directly from the \texttt{Qwen2.5-7B-Base} and \texttt{Qwen2.5-32B-Base} models using RL on each of these single-domain subsets and the combined mixed dataset. The detailed RL training setting is described in \Cref{sec:rl_setting}. \Cref{fig:cross_domain_heatmap} illustrates the cross-domain generalization performance trained on every single domain and mixed (rows) and evaluated on various target domain tasks (columns). For each training, we perform online test on 13 tasks for every 10 steps and select the checkpoint with the best average performance. \texttt{Qwen2.5-32B} exhibits heatmap patterns highly similar to those of \texttt{Qwen2.5-7B}; thus, its results are presented separately in the appendix \Cref{fig:cross_32B} for reference. 
To mitigate potential confounding effects from the instruction-following gap~\citep{zeng2025simplerl}, we confirm the following conditions: (1) each training run continues until validation accuracy plateaus, ensuring the initial instruction alignment phase is completed; (2) experiments are conducted on both 7B and 32B models, where the 32B model has stronger zero-shot instruction-following capabilities, and consistent results are observed across both; and (3) all runs use the same system prompt, with reward computation explicitly designed to be independent of special tokens such as \texttt{<think>} included in the prompt.
Our analysis yields several key findings below.

\subsection{Differential Transferability}
\label{sec:cross_domain_transfer}

\textbf{Domain identity matters.}  
\label{sec:transfer}
Math, Code, and Science consistently achieve stronger performance gains than other domains from cross-domain RL training~(\Cref{fig:cross_domain_heatmap}). We hypothesize that this asymmetry arises from the base model's extensive exposure to these domains during (continual-)pretraining~\citep{liu2024deepseek,yang2024qwen2}. The model thus contains deep knowledge in these domains which is effectively elicited and refined through RL, even when the training data originates from other domains. Our observation is consistent with the recent finding from \citet{shao2025spurious} that shows math reasoning can be easily triggered even with indirect signals like spurious rewards. In contrast, domains less represented in pretraining, such as Logic, Simulation, and Tabular, show limited improvement under cross-domain training. These results underscore the need to include underrepresented domains when curating RL training data, in order to build models with robust, general reasoning capabilities.

\textbf{Task difficulties matter.} As shown in \Cref{fig:cross_domain_heatmap}, easier tasks within Math (e.g., MATH500~\citep{hendrycks2021measuring}, AMC) and Code (e.g., HumanEval~\citep{chen2021evaluating}, MBPP~\citep{austin2021program}) readily exhibit positive transfer from other domains. In contrast, performance on more challenging benchmarks in those same domains, namely AIME24 and LiveCodeBench~\citep{jain2024livecodebench}, show considerably less improvement from cross-domain training. For the most challenging tasks, those with the lowest baseline absolute scores (e.g., ARC-AGI~\citep{chollet2024arc}, CodeI/O~\citep{li2025codei}), we observe marginal to negligible gains from training on non-native domain data. This variance in reasoning transferability suggests that cross-domain RL alone is not sufficient to effectively develop models with general, complex reasoning capability. Thus, advanced competence in multiple domains is dependent on training with domain-specific examples or with substantially more diverse data, perhaps drawing from multiple difficult domains.

\textbf{Mixed-domain training matches or exceeds single-domain performance.} RL training on a uniformly mixed dataset of reasoning tasks proves remarkably effective as shown in the last row of \Cref{fig:cross_domain_heatmap}. Across individual downstream benchmarks, models trained on this combined data achieve performance levels consistently comparable to, and sometimes surpassing, those attained by models trained exclusively on data from the target domain. This demonstrates that even a simple uniformly mixed dataset across multiple domains can significantly enhance general reasoning capabilities with minimal apparent interference between the six domains involved. However, future research should investigate whether interference remains negligible as the number and diversity of included domains is increased. Scaling in this manner might necessitate more refined domain balancing strategies.

\subsection{Reward and Response‑Length Dynamics}\label{subsec:rrld}

\begin{figure}[!t]
    \centering
    \includegraphics[width=\linewidth]{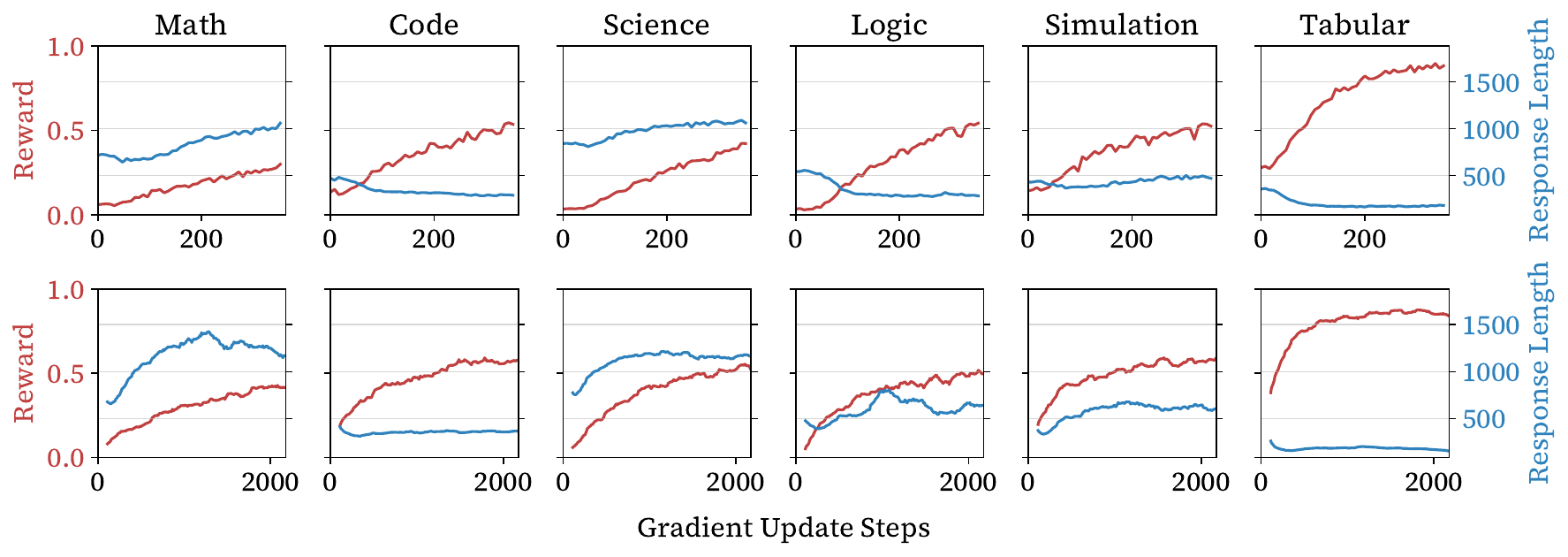}
    \caption{The reward and response length of each domain during RL training with: ({\bf top row}) single domain data (3k examples each) from \ours-18k; and ({\bf bottom row}): using the full \ours-18k mixture dataset. The x-axis is the number of gradient update steps.}
    \label{fig:dynamic}
\vspace{-5.0mm}
\end{figure}

\Cref{fig:dynamic} shows the reward and average response length across six reasoning domains during RL fine‑tuning. The upper row indicates RL training with 3k single-domain examples, while the lower row indicates joint training with 18k examples. The x‑axes are aligned so that each horizontal position represents an equal number of target domain samples.

In single-domain training, contrary to the common belief that RL drives models to produce longer responses, we observe strong domain effects: on Code, Logic, and Tabular tasks, the policy actually contracts its outputs, while Science and Math become more verbose, and Simulation remains largely unchanged. When we switch from single‑domain training to joint training on the full \ours‑18k mixture, rewards climb steeply within the first few hundred gradient steps across all domains. This demonstrates positive cross‑domain transfer. Joint training may also reshape length dynamics: for Logic, the single‑domain run monotonically shortens answers, but the multi‑domain run first lengthens and then shortens them, suggesting that shared representations learned from other domains modulate brevity preferences. In contrast, domains such as Tabular preserve their ``short‑answer'' tendency even under joint training, indicating that some length priors remain robust to cross‑domain influence.

\begin{table}[!t]
  \centering
  \vspace{3pt}
  \caption{Performance (best validation accuracy within 200 RL steps) comparison trained on difficulty-filtered (harder) v.s. unfiltered (easier) math data. Training on harder data improves in-domain performance but leads to degradation on easier cross-domain tasks (e.g., HumanEval and HiTab).}
    \resizebox{0.95\textwidth}{!}{
    \begin{NiceTabular}{lccc|cccc}
    \toprule
          & \multicolumn{3}{c}{\textbf{Math (in-domain)}} & \multicolumn{4}{c}{\textbf{Code \& Tabular (cross-domain)}} \\
                    \cmidrule(lr){2-8} 

          & \multicolumn{1}{l}{\textbf{MATH500}} & \multicolumn{1}{l}{\textbf{AMC}} & \multicolumn{1}{l}{\textbf{AIME24}} & \multicolumn{1}{l}{\textbf{HumanEval}} & \multicolumn{1}{l}{\textbf{LiveCodeBench}} & \multicolumn{1}{l}{\textbf{HiTab}} & \multicolumn{1}{l}{\textbf{Multihiertt}} \\
    \midrule
    \textbf{Unfiltered math} & 75.8  & 52.1  & 15.8  & 82.3  & 11.1  & 56.5  & 32.0 \\
    \textbf{Difficulty-filtered math} & 78.6  & 58.4  & 21.7  & 73.1  & 10.7  & 53.5  & 35.5 \\
    \textbf{$\Delta$ (+/-)} & \cellcolor[rgb]{ .988,  .894,  .839}+2.8 & \cellcolor[rgb]{ .973,  .796,  .678}+6.3 & \cellcolor[rgb]{ .973,  .796,  .678}+5.9 & \cellcolor[rgb]{ .557,  .663,  .859}-9.2 & \cellcolor[rgb]{ .906,  .902,  .902}-0.4 & \cellcolor[rgb]{ .706,  .776,  .906}-3.0 & \cellcolor[rgb]{ .988,  .894,  .839}+3.5 \\
    \bottomrule
    \end{NiceTabular}%
    }
  \label{tab:difficulty_filter_results}%
\end{table}%

\subsection{Effects of Training Data Difficulty}\label{subsec:etdd}

We further conducted an ablation to investigate the impact of training data difficulty on both in-domain performance and cross-domain transferability. Specifically, we trained the base \texttt{Qwen2.5-7B-Base} model using RL for 200 steps on either the complete, unfiltered Math domain data (representing a mix of difficulties) or a difficulty-filtered subset containing primarily harder Math problems from \ours. Other training configurations followed \Cref{sec:cross_domain_transfer}. We then evaluated performance on the Math tasks (in-domain) and selected Code and Tabular analysis tasks (cross-domain).

As shown in \Cref{tab:difficulty_filter_results}, training on difficulty-filtered math data consistently improves in-domain performance compared to unfiltered data, particularly on harder Math tasks like AMC (+6.3) and AIME24 (+5.9). However, the effects on cross-domain tasks are difficulty-dependent. Easier tasks such as HumanEval (-9.2) and HiTab (-3.0) suffer notable degradation. This decline was corroborated by observed accuracy collapse (large drops witnessed after approximately 100 steps and 150 steps respectively) during training on the filtered data. Conversely, performance on harder cross-domain tasks (LiveCodeBench, Multihiertt) shows minimal negative impact or positive change.

In summary, increasing training data difficulty within a domain can consistently enhance in-domain performance when using \texttt{Qwen2.5} as the base model. It is reasonable to perform aggressive difficulty filtering when the sole objective is maximizing performance within that specific domain on challenging tasks, e.g., AIME and LiveCodeBench.
While for cross-domain transfer, it introduces a risk of negative transfer to easier tasks in other domains. These results suggest that for beneficial cross-domain transfer, a more balanced distribution of training data difficulties, or the explicit inclusion of cross-domain data, may be more effective than solely increasing the difficulty of the source domain data.


\section{Main Experiment}\label{sec:mainexp}
Having established that RL behaviors vary significantly across domains~(\Cref{sec:analysis}), we now demonstrate the practical impact of multi-domain RL data through large-scale training experiments. We train 7B and 32B models on the complete \ours dataset to empirically validate that mixed-domain RL training enhances general reasoning capabilities at scale.
In this paper, we prioritize showing the core effectiveness of mixed-domain training, and we further study advanced data mixing and RL training strategies for future work.

\subsection{Experimental Setup}~\label{subsec.exp_setup}
\label{sec:rl_setting}

\begin{table}[!t]
  \centering
  \caption{Full 17 benchmark performance on 7B and 32B models. All baselines and our models are trained from \texttt{Qwen2.5-Base-7B/32B}. \ours shows significant improvements over baselines on general reasoning skills under two model sizes. $^\diamond$: ORZ represents Open-Reasoner-Zero~\citep{hu2025open}. $^\heartsuit$: Zebra Puzzle and LiveBench employ non-binary scoring that awards scores proportional to output correctness.
}
  \vspace{3pt}
\resizebox{0.95\textwidth}{!}{
    \begin{NiceTabular}{l|r|cccc|ccc}
    \toprule
    \multicolumn{2}{c|}{\textbf{\makecell{~~~~~~~~Models~$\rightarrow$\\Benchmarks~$\downarrow$}}} 
        & \cellcolor[rgb]{ .863,  .937,  1}\textbf{\makecell{\ours\\7B}}
        & \textbf{\makecell{General\\Reasoner\\7B}} 
        & \textbf{\makecell{ORZ\\7B$^\diamond$}} 
        & \textbf{\makecell{SimpleRL\\7B}} 
        & \cellcolor[rgb]{ .863,  .937,  1}\textbf{\makecell{\ours\\32B}} 
        & \textbf{\makecell{ORZ\\32B$^\diamond$}} 
        & \textbf{\makecell{SimpleRL\\32B}} \\
    \midrule
    \multicolumn{1}{c|}{\multirow{2}[1]{*}{\textbf{Math}}} & \multicolumn{1}{l|}{\text{AIME24\small{(avg@32)}}} & \cellcolor[rgb]{ .863,  .937,  1}17.50 & 17.08 & 16.25 & 15.60 & \cellcolor[rgb]{ .863,  .937,  1}34.89 & 47.50 & 27.20 \\
          & \multicolumn{1}{l|}{\text{MATH500}} & \cellcolor[rgb]{ .863,  .937,  1}77.25 & 70.40 & 80.80 & 87.00 & \cellcolor[rgb]{ .863,  .937,  1}86.00 & 89.80 & 89.60 \\
    \midrule
    \multicolumn{1}{c|}{\multirow{3}[1]{*}{\textbf{Code}}} & \multicolumn{1}{l|}{\text{LiveCodeBench\small{(avg@4)}}} & \cellcolor[rgb]{ .863,  .937,  1}16.49 & 8.51  & 5.47 & 6.72  & \cellcolor[rgb]{ .863,  .937,  1}29.30 & 22.04 & 19.80 \\
          & \multicolumn{1}{l|}{\text{HumanEval\small{(avg@4)}}} & \cellcolor[rgb]{ .863,  .937,  1}82.62 & 61.12 & 67.38 & 58.08 & \cellcolor[rgb]{ .863,  .937,  1}90.85 & 84.30 & 81.25 \\
          & \multicolumn{1}{l|}{\text{MBPP}} & \cellcolor[rgb]{ .863,  .937,  1}70.00 & 39.80 & 48.40 & 49.60 & \cellcolor[rgb]{ .863,  .937,  1}78.80 & 74.20 & 76.75 \\
    \midrule
    \multicolumn{1}{c|}{\multirow{2}[1]{*}{\textbf{Science}}} & \multicolumn{1}{l|}{\text{GPQA-diamond\small{(avg@4)}}} & \cellcolor[rgb]{ .863,  .937,  1}40.78 & 38.64 & 37.63 & 35.98 & \cellcolor[rgb]{ .863,  .937,  1}50.63 & 55.67 & 46.46 \\
          & \multicolumn{1}{l|}{\text{SuperGPQA}} & \cellcolor[rgb]{ .863,  .937,  1}31.80 & 30.64 & 29.75 &   27.29    & \cellcolor[rgb]{ .863,  .937,  1}43.60 & 46.05 & 37.73 \\
    \midrule
    \multicolumn{1}{c|}{\multirow{2}[1]{*}{\textbf{Logic}}} 
          & \multicolumn{1}{l|}{\text{ARC-AGI\small{(avg@4)}}} & \cellcolor[rgb]{ .863,  .937,  1}3.31 & 0.75  & 0.00  & 0.50  & \cellcolor[rgb]{ .863,  .937,  1}7.63 &  2.31     & 5.25 \\
          & \multicolumn{1}{l|}{\text{Zebra Puzzle$^\heartsuit$\small{(avg@4)}}} & \cellcolor[rgb]{ .863,  .937,  1}39.40 & 0.07  & 1.00  & 0.62  & \cellcolor[rgb]{ .863,  .937,  1}45.21 & 0.54  & 1.16 \\
    \midrule
    \multicolumn{1}{c|}{\multirow{3}[1]{*}{\textbf{Simulation}}} & \multicolumn{1}{l|}{\text{CodeI/O\small{(avg@4)}}} & \cellcolor[rgb]{ .863,  .937,  1}15.63 & 7.13  & 5.13  & 6.63  & \cellcolor[rgb]{ .863,  .937,  1}12.63 & 3.75  & 9.75 \\
          & \multicolumn{1}{l|}{\text{CruxEval-I}} & \cellcolor[rgb]{ .863,  .937,  1}61.72 & 63.63 & 69.38 & 56.25 & \cellcolor[rgb]{ .863,  .937,  1}80.63 & 71.13 & 72.63 \\
          & \multicolumn{1}{l|}{\text{CruxEval-O}} & \cellcolor[rgb]{ .863,  .937,  1}71.28 & 56.50 & 65.88 & 58.31 & \cellcolor[rgb]{ .863,  .937,  1}88.75 & 82.38 & 67.75 \\
    \midrule
    \multicolumn{1}{c|}{\multirow{3}[1]{*}{\textbf{Tabular}}} & \multicolumn{1}{l|}{\text{FinQA}} & \cellcolor[rgb]{ .863,  .937,  1}34.70 & 34.33 & 37.60 & 35.10 & \cellcolor[rgb]{ .863,  .937,  1}46.14 & 45.20 & 45.41 \\
          & \multicolumn{1}{l|}{\text{HiTab}} & \cellcolor[rgb]{ .863,  .937,  1}74.20 & 54.40 & 54.10 & 50.40 & \cellcolor[rgb]{ .863,  .937,  1}82.00 & 63.30 & 69.00 \\
          & \multicolumn{1}{l|}{\text{MultiHiertt\small{(avg@4)}}} & \cellcolor[rgb]{ .863,  .937,  1}44.94 & 31.62 & 38.10 & 37.57 & \cellcolor[rgb]{ .863,  .937,  1}55.28 & 52.83 & 52.83 \\
    \midrule
    \multicolumn{1}{c|}{\multirow{2}[1]{*}{\textbf{Others}}} & \multicolumn{1}{l|}{\text{IFEval}} & \cellcolor[rgb]{ .863,  .937,  1}35.81 & 39.56 & 32.72 & 36.69 & \cellcolor[rgb]{ .863,  .937,  1}55.45 & 38.26 & 55.27 \\
          & \multicolumn{1}{l|}{\text{LiveBench}$^\heartsuit$} & \cellcolor[rgb]{ .863,  .937,  1}18.57 & 19.76 & 12.64 & 15.20 & \cellcolor[rgb]{ .863,  .937,  1}34.30 & 28.78 & 28.33 \\
    \midrule
    \multicolumn{2}{c|}{\textbf{Average Score}}  & \cellcolor[rgb]{ .863,  .937,  1}\textbf{43.29} & 33.76 & 35.42 & 33.97 & \cellcolor[rgb]{ .863,  .937,  1}\textbf{54.24} & 47.53 & 46.25 \\
    \bottomrule
    \end{NiceTabular}%
}
\vspace{-3.5mm}
  \label{tab:leaderboard}%
\end{table}%

\paragraph{Training Configurations}

We use \texttt{verl}~\citep{sheng2024hybridflow} as the RL training framework and GRPO~\citep{shao2024deepseekmath} as the RL training algorithm. For general training hyper-parameters, we utilize the AdamW optimizer~\citep{loshchilov2017decoupled} with learning rate $1\times10^{-6}$, incorporating a linear warm-up of 10 RL steps. The prompt batch size is 512 for one RL step, and we sample 16 responses for each prompt with a temperature of 1.0. 
The mini-batch size is set to 64, i.e., 8 gradient updates for each RL step. 
The maximum number of tokens is 4k for input prompt and 8k for generation. We do not apply any KL or entropy loss in the training. The clipping parameter $\epsilon$ for GRPO is set to $0.2$.
Experiments are conducted on 20 GPU nodes, each equipped with 8 Hopper GPUs, for both RL training and evaluation. The 7B model training (3 epochs) and 32B model training (2 epochs) each required approximately 3 days of wall-clock time.

\paragraph{Baselines}  
For baseline comparisons, we compare the most performant RL-trained reasoning models with open data in general and math reasoning at 7B and 32B scale:
(1) \text{General Reasoner}~\citep{generalreasoner},
(2) \text{Open-Reasoner-Zero (ORZ)}~\citep{hu2025open}, and  
(3) \text{SimpleRL-Zoo}~\citep{zeng2025simplerl}
To ensure a fair comparison, we exclude baselines that use SFT-trained models or RL training on top of SFT models. All baselines and our models are directly RL-trained from \texttt{Qwen2.5}~\citep{yang2024qwen2} series base models.
It also allows us to directly evaluate the effectiveness of our \ours dataset without confounding factors introduced in SFT data. All benchmark scores reported in the following sections are evaluated using \textit{identical evaluation code and generation parameters} for both our models and baselines.

\paragraph{Evaluation Suite}
We build a comprehensive evaluation suite using both offline benchmarking for holistic assessment and online evaluation for monitoring training progress. The comprehensive evaluation dataset statistics are provided in \Cref{sec:eval_stats}.

\textit{Offline Evaluation.} Our offline suite encompasses 17 benchmarks across six domains and one unseen generalization, designed to rigorously assess reasoning capability through diverse and challenging tasks:
\begin{itemize}
    \item \textbf{Math}: AIME24~\citep{AIME2024} and MATH500~\citep{hendrycks2021measuring} for mathematical reasoning problems
    \item \textbf{Code}: HumanEval~\citep{chen2021evaluating}, MBPP~\citep{austin2021program}, and LiveCodeBench~\citep{jain2024livecodebench} for programming capabilities spanning basic generation to competitive programming
    \item \textbf{Science}: GPQA~\citep{gpqa} and SuperGPQA~\citep{supergpqa} covering graduate-level questions
    \item \textbf{Logic}: ARC-AGI~\citep{chollet2024arc} for abstract reasoning and Zebra Puzzle~\citep{logiczebra}\footnote{We generate a new test set for evaluation rather than using the original test set from \citet{logiczebra}.}
    for structured logical deduction
    \item \textbf{Simulation}: CodeI/O~\citep{li2025codei} and CRUXEval~\citep{gu2024cruxeval} for execution tracing abilities
    \item \textbf{Tabular}: HiTab~\citep{cheng2021hitab}, MultiHiertt~\citep{zhao2022multihiertt}, and FinQA~\citep{finqa} for heterogeneous and complex structured data reasoning
    \item \textbf{Others}: IFEval~\citep{zhou2023instruction} and LiveBench~\citep{white2024livebench} for unseen domain adaptation
\end{itemize}

For offline evaluation inference, we use \texttt{vLLM}~\citep{vllm_engine} framework with \texttt{temperature=1.0} and \texttt{top-p=0.7} for all the tasks and models. To ensure statistical reliability, we generate 4 samples per question for datasets with less than 400 samples and report averaged results. For AIME24, we generate 32 samples per question. Large datasets (more than 1000 samples) are subsampled to 1000 examples for efficiency.

\textit{Online Evaluation.} During RL training, we monitor progress using 13 signal tasks with up to 200 samples per task for computational efficiency. We select final checkpoints based on average performance across all online tasks. Online inference uses \texttt{temperature=1.0} and \texttt{top-p=1.0}, with 8 samples per AIME24 problem and 1 sample for other tasks.

\subsection{Results}\label{subsec.exp_results}
As shown in \Cref{tab:leaderboard}, \ours models demonstrate superior and well-balanced performance across all six evaluated domains. \ours-7B and \ours-32B achieve 43.29\% and 54.24\% respectively, outperforming the best baselines by 9.0\% and 6.7\%.

Domain-specific baselines exhibit significant weaknesses outside their specialization: math-focused models (Open-Reasoner-Zero, SimpleRL) and STEM-focused models (General-Reasoner) perform poorly on coding and logic tasks. Notably, Zebra Puzzle—a challenging synthetic constraint satisfaction task—shows substantial improvement from RL. \ours also maintains competitive results on unseen tasks (IFEval, LiveBench), demonstrating that multi-domain RL training effectively builds robust general reasoning capabilities.

\subsection{Analysis of Pass@\emph{k} Curves}


\label{subsec:passk_analysis}
Pass@\emph{k} measures the chance that at least one of k independent samples from a model is correct. Unlike accuracy, Pass@\emph{k} probes the reachable reasoning space of a model rather than its average-case behavior. Recent work \citep{yue2025does} reports that RL fails to improve Pass@\emph{k} at large k, concluding that the current RL recipe does not uncover fundamentally new reasoning capabilities beyond the base model. We revisit this claim using our Guru-7B and Guru-32B models on a diverse suite of tasks. The results reveal that the effect of RLVR on Pass@\emph{k} depends on several factors, with three key observations below.

\begin{figure*}[!htp]
\centering
\begin{subfigure}[t]{0.34\textwidth}
        \centering
        \includegraphics[width=\linewidth]{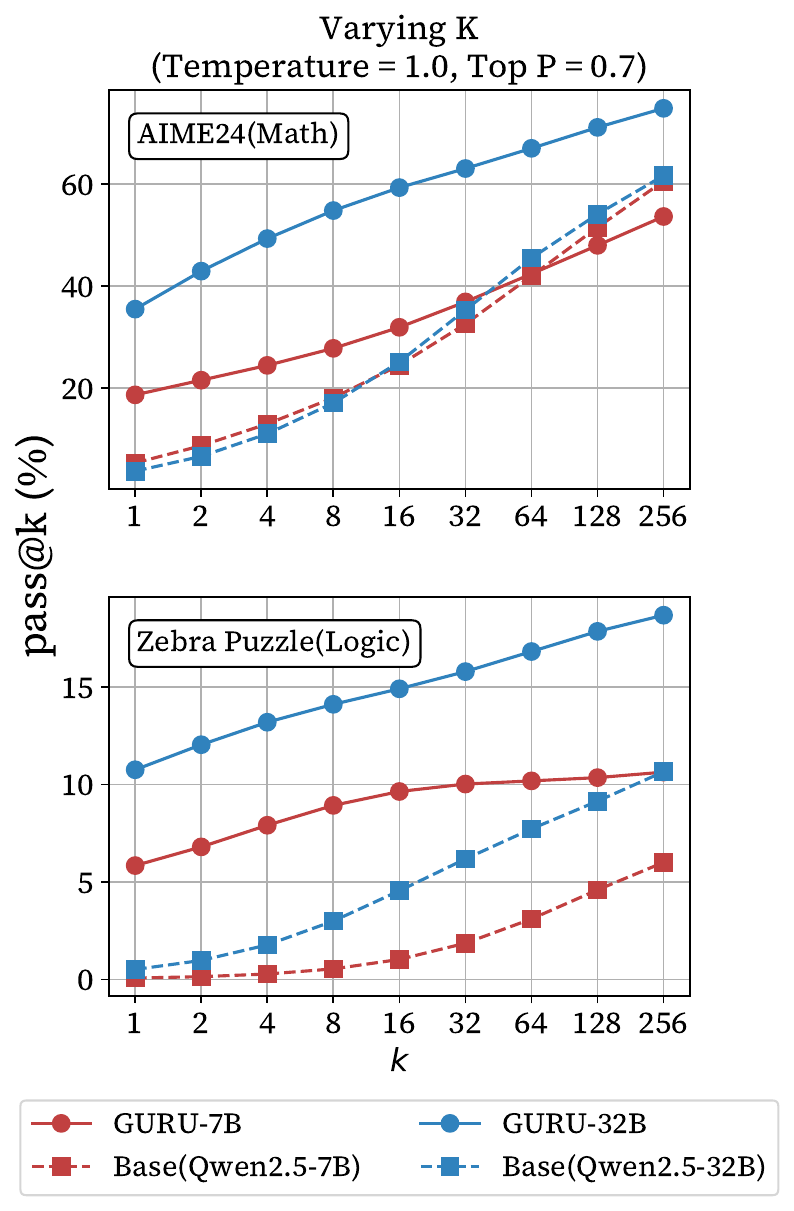}
        \caption{Varying $k$.}
        \label{fig:pass_k}
    \end{subfigure}
    \hfill
    \begin{subfigure}[t]{0.65\textwidth}
        \centering
        \includegraphics[width=\linewidth]{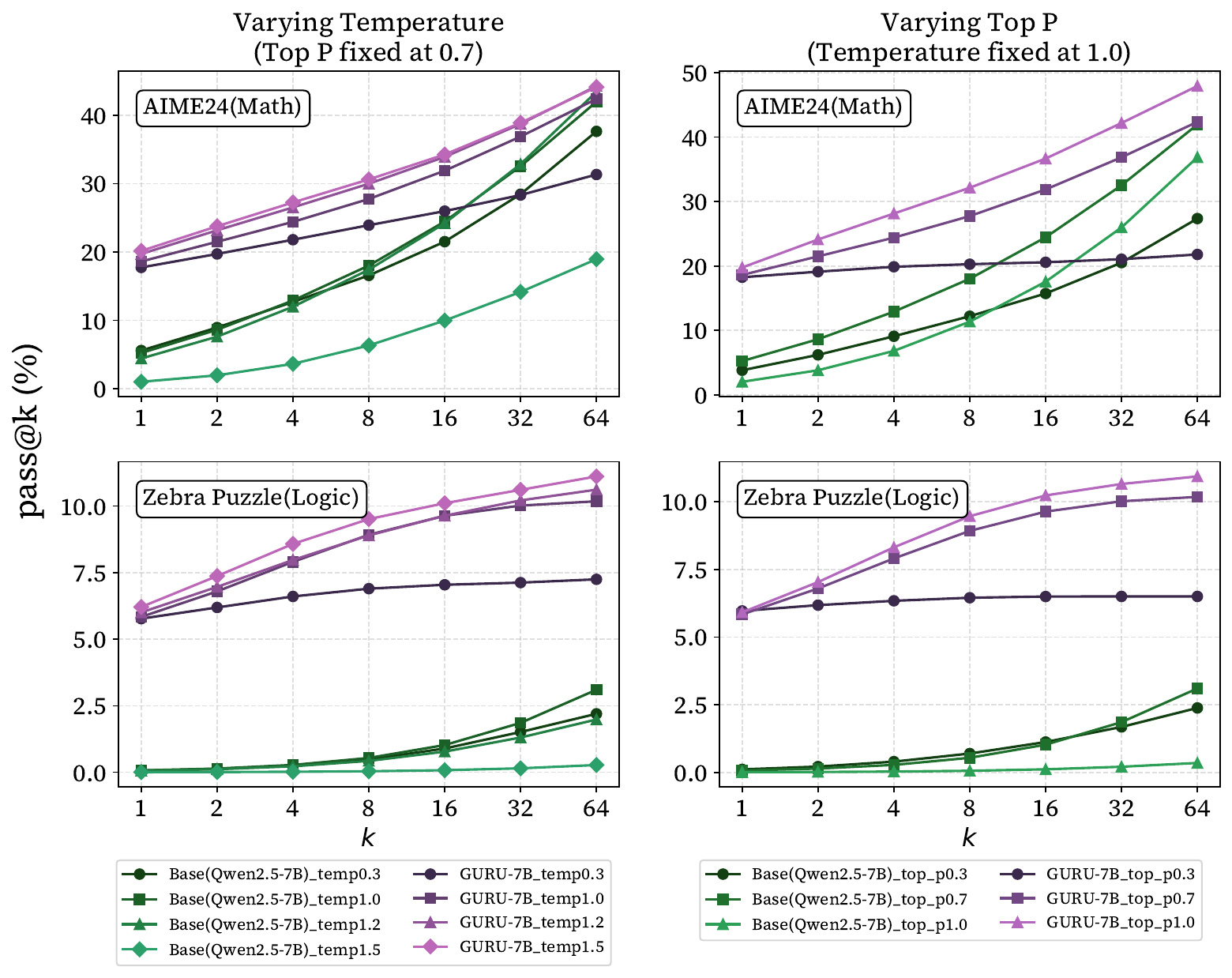}
        \caption{Varying temperature and top P.}
        \label{fig:temp_top}
    \end{subfigure}
    \caption{Pass@\emph{k} analysis of \ours. \textbf{(a)} Pass@\emph{k} on AIME24 (Math) and Zebra Puzzle (Logic) for base Qwen2.5-7B/32B vs. RL-tuned \ours-7B/32B. \textbf{(b)} Pass@\emph{k} under different decoding settings: higher sampling temperature or larger top-p broadens exploration and offsets RL-induced entropy collapse.} 
\label{fig:analyze_pass_k}
\end{figure*}

\textbf{Task‐dependent pattern.} As shown in \autoref{fig:pass_k}, Pass@\emph{k} curves exhibit different patterns across tasks rather than adhering to a uniform pattern. For example, on AIME, the curves of \ours-7B and the base model cross at $k=64$, reproducing the pattern reported in \citet{yue2025does}. Combined with our domain transfer study in \Cref{sec:transfer}, where math reasoning gains can be elicited via out-of-domain RL training, these results suggest that the observed improvements in math largely exploit the base model rather than introduce fundamentally new capabilities. However, this conclusion does not hold for every task. For example, on the synthetic Zebra Puzzle task, which is likely less exposed in the pretraining stage compared to math and coding, \ours-32B and \ours-7B both effectively expand the reasoning boundary of the base model. We also report the Pass@\emph{k} for checkpoints after different training steps in \autoref{sec:passk_dynamic}.

\textbf{Effect of model scale.}  
Even within a single task, Pass@\emph{k} behaviour changes with model scale. On AIME, the 32B curves never intersect, and RL consistently outperforms the base model. On the contrary, the 7B curves cross at $k{=}64$. On Zebra Puzzle, \ours-7B plateaus around $k{=}32$, while \ours-32B continues to rise across the entire sampled range. These patterns suggest that a stronger base model might more easily uncover new reasoning trajectories during RL finetuning, consistent with the DeepSeek-R1 study \citep{guo2025deepseekr1}, which claims RLVR works better for larger models.

\textbf{Influence of sampling hyperparameters.}
Our analysis of training logs reveals that the collapse in output entropy during RL training likely underlies the narrowing of Pass@\emph{k} at larger $k$. We find that adjusting decoding parameters can counteract this narrowing to a certain degree. Specifically, we explored raising the sampling temperature and tuning the nucleus‐sampling threshold (top‐$p$). As shown in \autoref{fig:temp_top}, compared to the standard decoding hyperparameter we applied, these settings broaden the model’s exploration, thereby improving Pass@\emph{k} at larger $k$. Given its sensitivity to decoding hyperparameters, we suggest treating Pass@\emph{k} as an informative but not exhaustive indicator of a model’s reasoning boundary and interpreting it with caution.

\section{Related Work}
\paragraph{Improving LLM Reasoning with RL}

Enhancing LLM reasoning with RL
has gained significant interest~\citep{shao2024deepseekmath, yuan2024advancing, pang2024iterative,  wang2024offline, havrilla2024teaching} due to its potential for self-improvement without relying on human-labeled solutions.
Recent milestones—most notably OpenAI o1 \citep{OpenAIo1} and DeepSeek-R1 \citep{guo2025deepseekr1}—demonstrate that RL can be scaled to industry-level models, markedly advancing state-of-the-art reasoning benchmarks and revealing emergent capabilities such as long reasoning chains. Following initial successes, a significant body of work has explored open recipes to build strong reasoning models. Efforts such as Open-Reasoner-Zero~\citep{hu2025open}, Skywork-OR1~\citep{skywork-or1-2025}, DeepScaler~\citep{deepscaler2025}, and SimpleRL~\citep{zeng2025simplerl} have notably leveraged extensive mathematical data to achieve state-of-the-art performance on complex math benchmarks. Similarly, DeepCoder~\citep{deepcoder2025} focused on RL for code generation tasks. While powerful within their specific areas, this domain-specific focus inherently limits the generalizability of the resulting models across the broader landscape of reasoning tasks. Concurrent works like General-Reasoner~\citep{generalreasoner} and Nemotron-CrossThinker~\citep{akter2025nemotron} have begun to explore broader domains for RL training. However, their scope remains constrained to STEM problems, leaving many crucial scenarios, including comprehensive coding, logic, and tabular analysis unexplored by large-scale open RL efforts. Addressing this critical gap in open resources and models for general reasoning via RLVR, we introduce \ours, a novel multi-domain dataset spanning six reasoning domains: Math, Code, Science, Logic, Simulation, and Tabular. Utilizing \ours, we train \ours-7B/32B, general reasoning models optimized via RL exclusively on open data using our \ours dataset and see state-of-the-art results on general reasoning among open models trained with publicly available data using RL.

\paragraph{Understanding RL for Reasoning}
Alongside the development of specialized models, a parallel line of research has critically examined the underlying mechanisms of how reinforcement learning enhances reasoning. This inquiry has revealed that RL for reasoning may not fundamentally lead to the acquisition of new reasoning capabilities, but rather serves as an elicitor of latent abilities already present in the base model. For instance, \citet{yue2025does} find that RL primarily improves the sampling efficiency of correct reasoning paths that already exist in the base model's distribution, rather than expanding its core reasoning capacity. This amplifier hypothesis is further substantiated by several surprising findings. Research has shown that significant performance gains can be triggered by seemingly uninformative signals, such as rewards that are random or even known to be incorrect \citep{shao2025spurious}, or through training on just a single example \citep{one-shot-rl}. In an extreme case, \citet{min-entropy} demonstrate that strong reasoning performance can be unlocked without any labeled data at all, using only entropy minimization as the training objective. However, this body of work also highlights that these effects are highly model-dependent. The success of such techniques is often pronounced in models like Qwen, while being significantly less effective on other model families \citep{shao2025spurious, zeng2025simplerl, min-entropy}. Recent work has begun to challenge this amplifier-only perspective: \citet{prolongRL} demonstrate that prolonged RL training can discover novel reasoning strategies inaccessible to base models even under extensive sampling, particularly in domains where the base model initially struggles. However, our work systematically revisits these underlying mechanisms from a novel cross-domain perspective and at multiple scales, attempting to explore whether the role of RL exhibits high domain-dependency and whether synergistic multi-domain training can yield mutual benefits for general reasoning.


\section{Conclusion}
\label{sec:conclusion}

In this work, we introduce \ours, a curated reinforcement learning (RL) dataset of 92K high-quality examples spanning six diverse domains—Math, Code, Science, Logic, Simulation, and Tabular reasoning—designed to bridge a critical gap in RL-based reasoning research. Our extensive empirical analysis reveals that the effects of RL on reasoning behaviors are highly domain-sensitive. For example, while RL can effectively elicit pretrained knowledge in well-covered domains like Math, Code, and Science, it is more likely to foster genuine reasoning capabilities in underrepresented domains in pretraining. We develop \ours-7B and \ours-32B, two general-purpose reasoning models trained via RL on the full \ours dataset. Across 17 benchmark tasks across six reasoning domains, these models achieve state-of-the-art performance among open models trained with RL on open data, marking a substantial advance in general reasoning capability. We release the full \ours dataset, models, evaluation suite, and training code to encourage further research into general-purpose RL for reasoning.

\newpage
\bibliographystyle{plainnat}
\bibliography{references}

\newpage
\beginappendix
\section{Limitations}
We are encouraged by the promising improvements observed on general reasoning datasets through the application of RL with our 92K sample data. However, for 7B and 32B models, our current training only lasts three and two epochs due to computing resource constraints. It is conceivable that extended RL training might yield even more pronounced results according to \citet{prolongRL} and \citet{yu2025dapo}.

Our approach initiated training directly from a base model using RL, bypassing an initial SFT phase. This decision was made to ensure that our conclusions regarding cross-domain generalization remained independent of specific SFT data dependencies. However, we expect that integrating an SFT then RL pipeline could further enhance performance largely, and we plan to explore this in future work.

\section{Training Data Curation Details}
\label{sec:data_appendix}
In this section, we elaborate the data sourcing and curation details of all used dataset in \ours.

\textbf{New Assets}: We are committed to full transparency and open science. All the new assets we've created, as well as those repurposed from existing datasets, are fully open-source and publicly available. We believe this approach fosters collaboration and accelerates progress within the research community.

\begin{table}[htbp]
\caption{Public source data links for all datasets used in each domain for \ours.}
\label{tab:data-sourcing}
\centering
\begin{tabular}{lll}
\toprule
\textbf{Domain} & \textbf{Dataset} & \textbf{Data Source Link} \\
\midrule
\multirow{3}{*}{\textbf{Math}}%
    & OR1                & \href{https://huggingface.co/datasets/Skywork/Skywork-OR1-RL-Data}{\hf\;Skywork/Skywork-OR1-RL-Data} \\
    & DAPO               & \href{https://huggingface.co/datasets/BytedTsinghua-SIA/DAPO-Math-17k}{\hf\;BytedTsinghua-SIA/DAPO-Math-17k} \\
    & DeepScaler         & \href{https://huggingface.co/datasets/agentica-org/DeepScaleR-Preview-Dataset}{\hf\;agentica-org/DeepScaleR-Preview-Dataset} \\
\midrule
\multirow{4}{*}{\textbf{Code}}%
    & LeetCode           & \href{https://huggingface.co/datasets/newfacade/LeetCodeDataset}{\hf\;newfacade/LeetCodeDataset} \\
    & TACO-Verified      & \href{https://huggingface.co/datasets/likaixin/TACO-verified}{\hf\;likaixin/TACO-verified} \\
    & PrimeIntellect     & \href{https://huggingface.co/datasets/agentica-org/DeepCoder-Preview-Dataset}{\hf\;agentica-org/DeepCoder-Preview-Dataset} \\
    & LiveCodeBench      & \href{https://huggingface.co/datasets/agentica-org/DeepCoder-Preview-Dataset}{\hf\;agentica-org/DeepCoder-Preview-Dataset} \\
\midrule
\textbf{Science}%
    & WebInstruct-Verified & \href{https://huggingface.co/datasets/TIGER-Lab/WebInstruct-verified}{\hf\;TIGER-Lab/WebInstruct-verified} \\
\midrule
\multirow{6}{*}{\textbf{Logic}}%
    & Zebra Puzzle       & -- \\
    & Ordering Puzzle    & -- \\
    & Graph Puzzle       & -- \\
    & ARC-AGI            & \href{https://github.com/fchollet/ARC-AGI}{\gh\;fchollet/ARC-AGI} \\
    & ARC-AGI-2          & \href{https://github.com/arcprize/ARC-AGI-2}{\gh\;arcprize/ARC-AGI-2} \\
    & BARC               & \href{https://huggingface.co/datasets/barc0/200k_HEAVY_gpt4o-description-gpt4omini-code_generated_problems}{\hf\;barc0/200k\_HEAVY\_gpt4o...problems} \\
\midrule
\textbf{Simulation}%
    & Code I/O (PyEdu)   & \href{https://huggingface.co/datasets/hkust-nlp/CodeIO-PyEdu-Reasoning-Raw}{\hf\;hkust-nlp/CodeIO-PyEdu-Reasoning-Raw} \\
\midrule
\multirow{2}{*}{\textbf{Tabular}}%
    & HiTab              & \href{https://github.com/microsoft/HiTab}{\gh\;microsoft/HiTab} \\
    & MultiHierTT        & \href{https://github.com/psunlpgroup/MultiHiertt}{\gh\;psunlpgroup/MultiHiertt} \\
\bottomrule
\end{tabular}
\end{table}


\begin{figure}[htbp]
    \centering
    \includegraphics[width=0.85\linewidth]{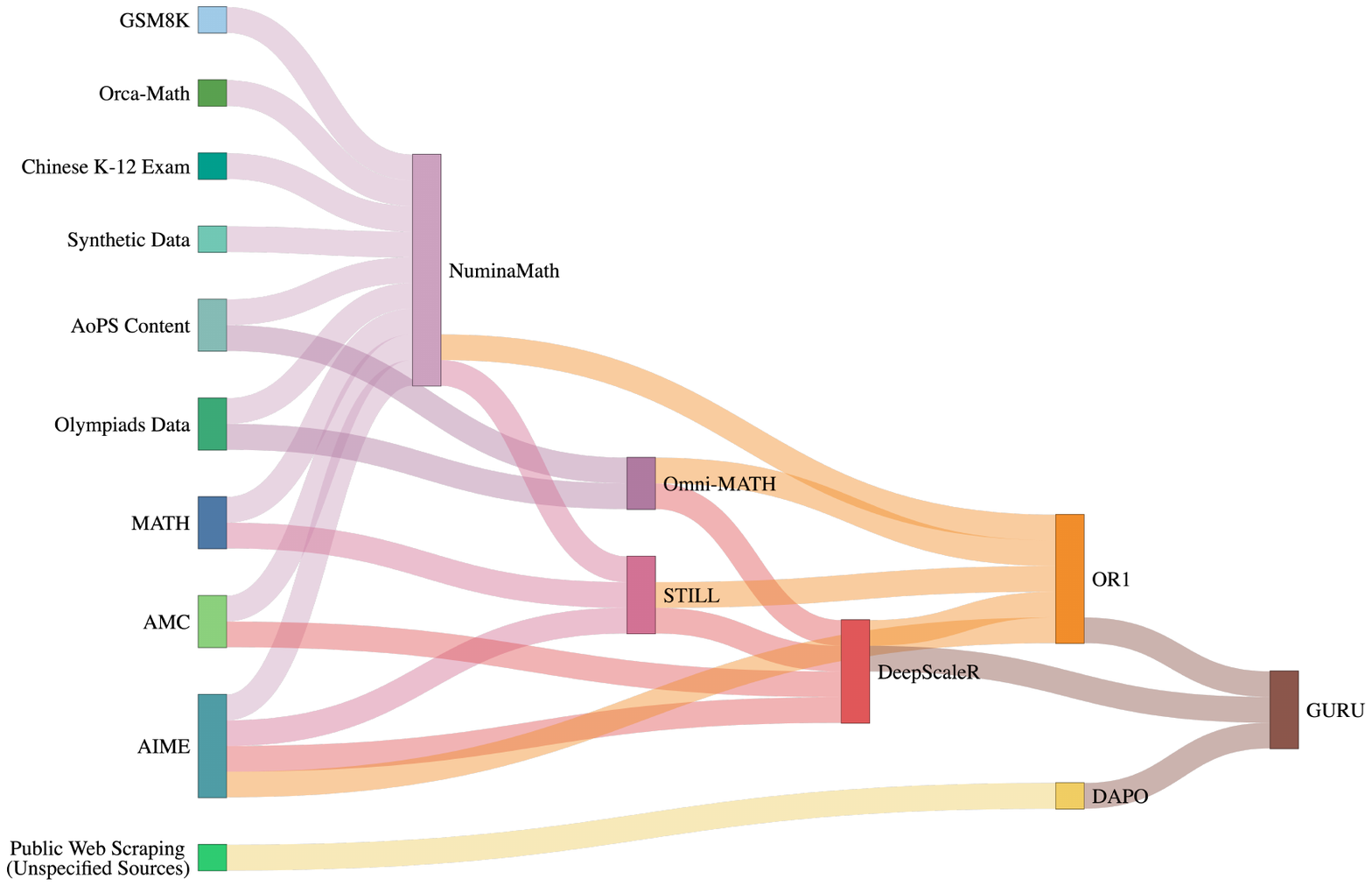}
    \caption{The sankey diagram tracing the provenance of all math data included in \ours.}
    \label{fig:sankey-math}
\end{figure}

\subsection{Math}
For our math subset, we focus on gathering some of the most effective datasets recently released. We prioritize OR1~\citep{skywork-or1-2025}, DAPO~\citep{yu2025dapo}, and DeepScaler~\citep{deepscaler2025} due to their strong performance in advancing mathematical reasoning in large language models. For other math reasoning datasets, such as BigMath~\citep{albalak2025bigmathlargescalehighqualitymath}, we observe that their quality is inconsistent, and they contain more duplicate queries, which led us to exclude them from our selection. We identify the original sources of these datasets and, as shown in \Cref{fig:sankey-math}, meticulously trace their origins to understand their provenance and potential overlaps. We further standardize the data formats across all selected datasets, ensuring consistency and ease of integration into our training pipeline. A critical step is identifying and eliminating overlapping or duplicate problems within and across datasets. This ensures that our model is exposed to a diverse set of unique challenges, preventing overfitting to specific problem types and promoting broader generalization. By carefully curating these datasets, we aim to provide a robust and varied foundation for training and evaluating advanced mathematical reasoning capabilities in LLMs.

We instruct the model to output its final answer in \texttt{\textbackslash boxed\{\}}, and extract the content in the box as the model's predicted answer. Following PRIME~\citep{cui2025process}, the reward is granted if the LLM's generated answer either normalizes to the same string or is numerically equal (convertible to floats and equivalent) to the ground truth, or if \texttt{Sympy}\footnote{\url{https://www.sympy.org/en/index.html}} can simplify the difference between the predicted and ground truth expressions to zero.

\begin{figure}[htbp]
    \centering
    \includegraphics[width=0.85\linewidth]{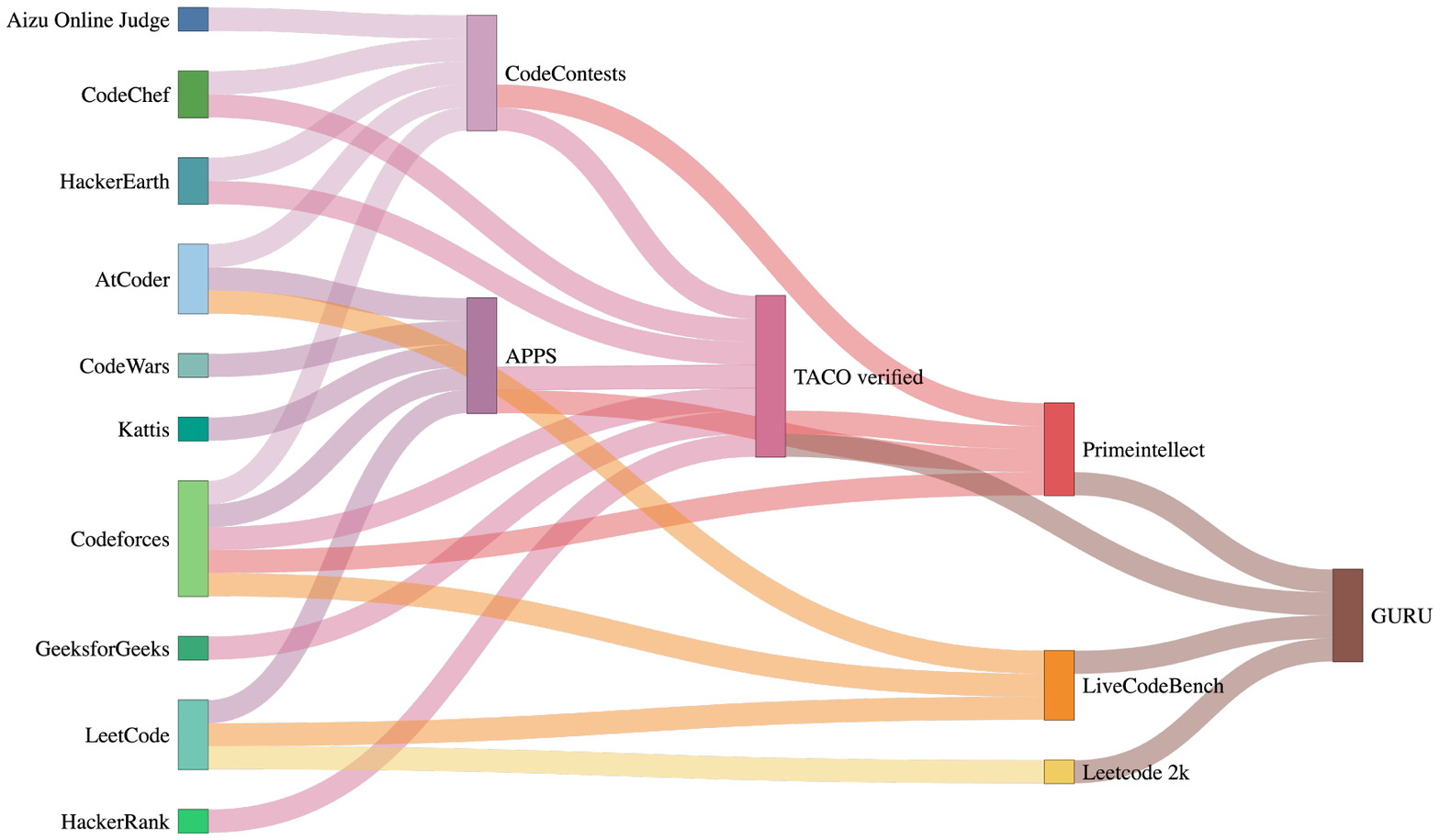}
    \caption{The sankey diagram tracing the provenance of all code data included in \ours.}
    \label{fig:sankey-code}
\end{figure}
\subsection{Code}

Our code subset consists of programming problems collected from a variety of sources, including online coding platforms and programming competitions. Specifically, we include real-world challenges from LeetCode~\citep{leetcode}, curated and validated tasks from TACO-Verified~\citep{likaixin2024taco-verified}, problems from PrimeIntellect~\citep{2025synthetic1}, and historical contest problems from LiveCodeBench~\citep{jain2024livecodebench}. For PrimeIntellect and LiveCodeBench, we adopt the pre-filtered subsets provided by the DeepCoder dataset~\citep{deepcoder2025}, which applies rigorous cleaning and verification protocols. A comprehensive breakdown of data source and corresponding Hugging Face links is provided in \Cref{tab:data-sourcing}. We also construct a Sankey diagram \Cref{fig:sankey-code} that traces the provenance of all code data included in our benchmark.

Each problem is presented as a program synthesis task requiring the model to generate a Python solution that satisfies a given problem specification to pass certain test cases. Specifically, we have two problem formats: (1) \textit{pure function format}, where a function is implemented and validated via inline \texttt{assert} statements; and (2) \textit{stdin/stdout format}, where the model must write a full script that reads from standard input and writes to standard output.
All programs are executed with a uniform 30-second timeout and 10 GB memory limit to ensure stability and computational efficiency. Programs that successfully pass all test cases receive a reward of 1. We discard any example whose reference solution fails its own unit tests. To ensure stable reward computation in high-concurrency training environments, we further filter out samples with large inputs (e.g., stdin exceeding 1MB) and randomly subsample at most 8 test cases per problem.
Currently, our setup only focuses on Python, we leave the inclusion of additional programming languages (e.g., Java, C++, Go, etc.) as future work, to explore both cross-lingual transfer and the necessity of language-level diversity in RL training for code.

\subsection{Science}
Our science data leverages WebInstruct-Verified~\citep{generalreasoner}. This dataset comprises over 230K verifiable question-answer pairs across diverse scientific disciplines, including physics, chemistry, biology, and finance. To ensure the dataset's suitability for reinforcement learning, we retain only university- and PhD-level questions in physics, chemistry, and biology, removing multiple-choice and boolean formats to avoid answer shortcuts. We also discard samples with extremely high numeric precision, which often lead to brittle or noisy reward evaluations.

We instruct the model to output its final answer in \texttt{\textbackslash boxed\{\}}, and extract the content in the box as the model's predicted answer. To calculate the reward, we apply a 1.5B model verifier\footnote{\url{https://huggingface.co/TIGER-Lab/general-verifier}} provided by \citet{generalreasoner} to compare the reference and model prediction.

\subsection{Logic}  
To evaluate and improve the logical reasoning capabilities of language models, we design three synthetic tasks and collect three existing datasets, all of which require multi-step deduction under structured constraints.
For all logical tasks, we instruct the model to wrap the prediction into special tags \texttt{<answer>} and \texttt{</answer>} and extract the prediction from there. The model prediction will be parsed into a list, and the reward is granted if every element matches the reference answer.

\paragraph{Zebra Puzzles}
Zebra puzzles are a class of logic grid problems that require deducing a complete set of attribute assignments (e.g., who owns the zebra) based on a set of relational constraints. The task becomes exponentially more difficult as the number of entities and constraints increases, making it a challenging benchmark for multi-step reasoning. We construct puzzles by first sampling a ground-truth table over randomly chosen attribute categories and values, then backsolving to generate constraints. These include positional relations (e.g., left/right, middle), equality/inequality constraints, disjunctive logic, relative distances, and parity-based rules. We define $20$ difficulty levels based on constraint complexity, support puzzles up to $15 \times 10$ in size, and optionally include redundant constraints to increase distractor robustness. To ensure a meaningful symbolic reasoning challenge, we retain only puzzles with grid sizes of at least $10$ objects and $5$ attributes.

\paragraph{Ordering Puzzles}
Ordering puzzles require models to recover a linear order over $n$ objects (up to $50$) based on a set of relational constraints. We first sample a ground-truth permutation and then generate constraints from it. These include absolute positions (e.g., ``A is at position 3''), relative positions (e.g., ``A is to the left of B''), adjacency relations (e.g., ``A is immediately to the right of B''), and distance-based constraints (e.g., ``There are 3 items between A and B''). To ensure uniqueness, we apply constraint propagation to select a minimal satisfiable subset. Our dataset spans diverse entity types, such as animals, people, and vehicles. To maintain task complexity, we filter out samples with fewer than $20$ objects.

\paragraph{Graph Puzzles}
Graph puzzles task models with identifying a valid path through a directed graph defined by natural language implications. We build upon the structure of PrOntoQA~\citep{saparov2023languagemodelsgreedyreasoners}, extending it to free-form language using the setup of Graph Search Tasks~\citep{saparov2025transformersstrugglelearnsearch}. Each instance defines a set of predicates that implicitly specify the graph structure, along with a query that requires multi-hop traversal. We vary the number of nodes, edge density, and path length to control task difficulty, with longer and denser paths posing greater challenges for compositional reasoning. At heuristc filtering stage, we retain only instances with more than $10$ nodes and at least $3$-hop queries to ensure the graph structure supports multi-step traversal reasoning.

\paragraph{ARC-AGI}
The original \textit{Abstraction and Reasoning Corpus for Artificial General Intelligence}~\citep{chollet2024arc} (ARC-AGI-1) comprises $400$ public training tasks and an equally sized public evaluation split, all expressed as grid-based input–output transformations. Tasks are deliberately designed to be straightforward for humans yet challenging for machines, probing fluid, compositional intelligence over a space of $30\times30$ coloured grids. Each instance supplies 2–5 demonstration pairs (as input examples) and one unseen test grid. We include the full training split of ARC-AGI-1 when constructing \ours.

\paragraph{ARC-AGI-2}
ARC-AGI-2~\citep{chollet2025arcagi2newchallengefrontier} expands to ARC-AGI to $1{,}000$ public training tasks and $120$ public evaluation tasks while introducing two hidden private test sets for competition use. The new version re-balances difficulty so that average human solvers achieve roughly $66\%$ accuracy, sharpens the trial budget to two attempts per test grid, and retains the canonical grid-world format. By fusing novel tasks with the ARC-AGI-1 corpus, ARC-AGI-2 covers a broader concept space (symmetry, physics-like dynamics, object counting, etc.) while preserving the ``few demonstrations, zero prior training'' ethos. We include the public training portion in \ours.

\paragraph{BARC}
BARC~\citep{li2024combininginductiontransductionabstract} is a large-scale synthetic extension containing approximately $200{,}000$ ARC-style problems generated from $160$ manually curated seed programs via an LLM-guided remix pipeline. Each synthetic task is accompanied by an executable Python reference code and an input generator, enabling controllable difficulty and unlimited fresh samples for augmentation or test-time fine-tuning.  We sample $3.4k$ examples whose input lengths do not exceed the token limit of our training.

\subsection{Simulation}

Simulation is the process of creating a model of a real or hypothetical system and executing it to observe its dynamic behavior over time. This process is inherently linked to reasoning because it requires the systematic application of logical inference to predict the consequences of actions or changes within a system. In this domain, our work focuses on the specific subproblem of \textit{simulating Python program execution}. Prior research has explored various facets of this problem. \citet{zaremba2014learning} trained LSTM to evaluate short computer programs. \citet{nye2021show} demonstrated the capability of LLMs to predict code execution traces. CruxEval~\citep{gu2024cruxeval} further advanced this by introducing a benchmark to evaluate LLM performance in both inferring inputs from given function outputs and predicting outputs from given inputs. Recent works, such as SemCoder~\citep{ding2024semcoder} and Code I/O~\citep{li2025codei}, have used SFT with code execution prediction tasks to enhance LLM capabilities in code generation and general reasoning.

In \ours, we repurpose the raw data from Code I/O for our RL training. The Code I/O dataset, sourced from PyEdu~\citep{smollm-corpus_dataset}, comprises 4.5 million verified programs with corresponding input-to-output and output-to-input samples. Inputs for this dataset were generated by DeepSeek-v2.5, and outputs were obtained through actual function execution. Since multiple input-output variants may exist for each program, to ensure diversity within our training data, we first applied a heuristic filter to retain only a single pair per problem, reducing the dataset to 227K programs. Subsequently, we randomly sampled 12K I/O pairs. A final difficulty-based filtering step was then applied, resulting in a refined dataset of 3.7K I/O samples used for our RL training. 

We instruct the model to wrap the answer in a json code block with Markdown grammar, and extract the content in the code block to compare with the reference answer. The reward will be granted if they are strictly the same.

\subsection{Tabular}

Tabular data are ubiquitous and serve as vital media for data organization and analysis. Table Question Answering (TQA), which involves deriving an answer from one or more tables given a natural language query, is a fundamental and critical problem in data understanding. Early work in TQA, such as WikiTQ~\citep{pasupat2015compositional} and HybridQA~\citep{chen2020hybridqa}, explored answering queries on WikiTables, sometimes incorporating accompanying text. The complexity of TQA increased with datasets like HiTab~\citep{cheng2021hitab} and MultiHiertt~\citep{zhao2022multihiertt}, which focused on hierarchical tables featuring multiple levels of headers, demanding more sophisticated reasoning. More specialized applications emerged with FinQA~\citep{chen2021finqa} and ConvFinQA~\citep{chen2022convfinqa}, addressing TQA within the financial domain.

In \ours, we adapt the training sets from HiTab (7.5K samples) and MultiHiertt (7.8K samples) for our RL training. We chose these datasets because they necessitate complex multi-hop reasoning over tabular structures. For table representation, we linearize the tables into a markdown format. If multiple tables are present for a single query, we concatenate them, separated by a line break.

We instruct the model to output its final answer in \texttt{\textbackslash boxed\{\}}, and extract the content in the box as the model's predicted answer. To calculate the reward, we apply the same program for math reasoning to compare the model prediction and the reference answer.

\section{Evaluation Benchmark Details}
\label{sec:eval_stats}
We provide detailed benchmark statistics in \Cref{tab:eval_stats}. For online evaluations, we randomly sample up to 200 examples from each dataset to ensure efficiency. An exception is ARC-AGI, where we select only those examples with input prompt lengths shorter than 4,096 tokens. This introduces a distributional discrepancy compared to the full offline set (400 examples), which partially explains the performance gap between online and offline settings. Another exception is MultiHiertt, where we specifically include only examples with table inputs and no accompanying textual context, to deliberately evaluate the model’s multi-table reasoning capabilities. For offline evaluations, large datasets (i.e., those with more than 1,000 examples) are subsampled to 1,000 for computational efficiency (SuperGPQA, HiTab). To ensure statistical reliability, we generate four completions per example for datasets with fewer than 400 samples and report the average results. For AIME24, we generate 32 completions per example. For LiveBench, we evaluate the language, reasoning, and data analysis subsets, following \citet{li2025codei}.

\begin{table}[!t]
\caption{Evaluation benchmark statistics across all domains. The left table presents offline evaluations, while the right table presents online evaluations. Benchmarks marked with * indicate that a subset of the official evaluation set was used, either for computational efficiency or to focus on specific reasoning capabilities.}
\centering
\label{tab:eval_stats_split}
\resizebox{\textwidth}{!}{%
\begin{minipage}[t]{0.49\textwidth}
\centering
\textbf{Offline Evaluation} \vspace{0.5em}

\begin{NiceTabular}{l|l|r}
\toprule
\textbf{Domain} & \textbf{Benchmark} & \textbf{Size} \\
\midrule
\multirow{2}{*}{Math}
  & AIME24 \small{(avg@32)} & 30 \\
  & MATH500         & 500 \\
\midrule
\multirow{3}{*}{Code}
  & LiveCodeBench v5 \small{(avg@4)} & 279 \\
  & MBPP             & 500 \\
  & HumanEval \small{(avg@4)}  & 164 \\
\midrule
\multirow{2}{*}{Science}
  & GPQA-diamond \small{(avg@4)}    & 198 \\
  & SuperGPQA*       & 1000 \\
\midrule
\multirow{2}{*}{Logic}
  & ARC-AGI \small{(avg@4)}      & 400 \\
  & Zebra Puzzle \small{(avg@4)} & 200 \\
\midrule
\multirow{3}{*}{Simulation}
  & Code I/O      & 200 \\
  & CruxEval-I    & 800 \\
  & CruxEval-O    & 800 \\
\midrule
\multirow{3}{*}{Tabular}
  & FinQA         & 1100 \\
  & HiTab* \small{(avg@4)}   & 1000 \\
  & MultiHiertt* \small{(avg@4)}  & 336 \\
\midrule
\multirow{2}{*}{Others}
  & IFEval        & 541 \\
  & LiveBench* \small{(avg@4)} & 440 \\
\bottomrule
\end{NiceTabular}
\end{minipage}%
\hfill
\begin{minipage}[t]{0.49\textwidth}
\centering
\textbf{Online Evaluation} \vspace{0.5em}

\begin{NiceTabular}{l|l|r}
\toprule
\textbf{Domain} & \textbf{Benchmark} & \textbf{Size} \\
\midrule
\multirow{2}{*}{Math}
  & AIME24 \small{(avg@8)} & 30 \\
  & AMC \small{(avg@4)}    & 332 \\
  & MATH500         & 500 \\
\midrule
\multirow{3}{*}{Code}
  & LiveCodeBench v5 & 279 \\
  & MBPP             & 500 \\
  & HumanEval        & 164 \\
\midrule
Science & SuperGPQA & 200 \\
\midrule
\multirow{3}{*}{Logic}
  & ARC-AGI  & 200 \\
  & Zebra Puzzle     & 200 \\
  & Ordering Puzzle  & 100 \\
\midrule
Simulation & Code I/O & 200 \\
\midrule
\multirow{2}{*}{Tabular}
  & HiTab         & 200 \\
  & MultiHiertt   & 200 \\
\bottomrule
\end{NiceTabular}
\end{minipage}%
\label{tab:eval_stats}
}
\end{table}

\newpage
\section{More Results}
\subsection{Cross-Domain Transferability}
\label{sec:cross_domain_32b}

\begin{figure}[ht]
    \centering
    \includegraphics[width=1.\linewidth]{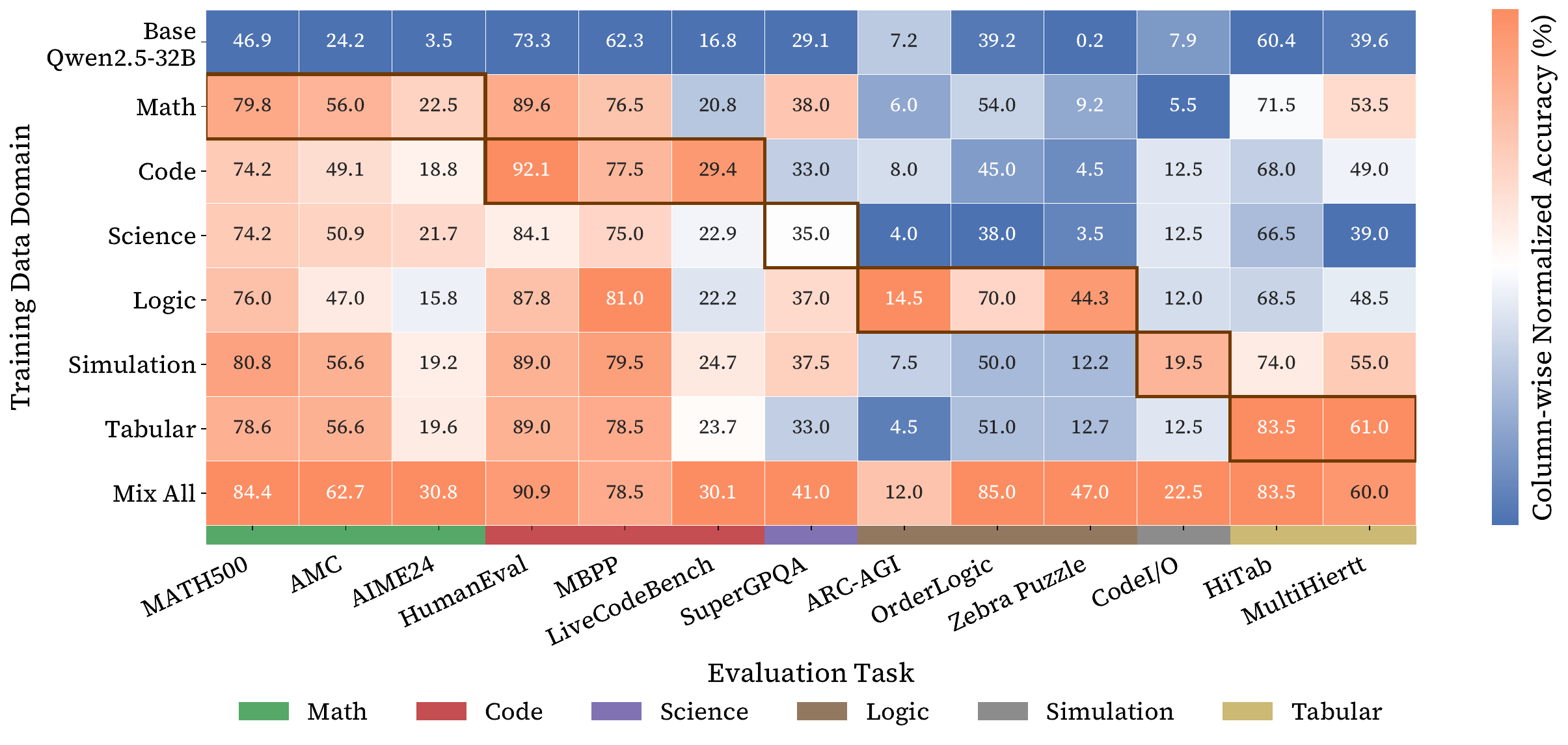}
    \caption{\textbf{Cross-Domain RL Transfer Performance per Task with \texttt{Qwen-32B-Base}.}
    This figure replicates the heatmap analysis from \Cref{fig:cross_domain_heatmap} using \texttt{Qwen-32B-Base}. Warmer colors indicate higher performance gains from RL training on each source domain (rows) when evaluated on target domains (columns). Each cell shows the min-max normalized validation accuracy, with the best per-domain accuracy (\%) labeled. In-domain training results are outlined.}
    \label{fig:cross_32B}
\end{figure}

To supplement our main findings on differential transferability (see \Cref{sec:cross_domain_transfer}), we also conduct the same set of cross-domain RL experiments using a larger backbone, \texttt{Qwen-32B-Base}. Results are presented in \Cref{fig:cross_32B}, following the same experimental setup and visualization protocol as described in \Cref{fig:cross_domain_heatmap}.

We observe that the overall pattern of cross-domain transferability in \texttt{Qwen-32B-Base} closely mirrors that of the 7B variant, reaffirming our core conclusions regarding the role of pretraining priors, task difficulty, and training composition. Specifically, we find that 
(1) domains with strong pretraining coverage—such as Math, Code, and Science—continue to exhibit substantial gains under cross-domain RL, suggesting that reinforcement learning primarily serves to elicit capabilities already encoded in the base model. In the 32B setting, these benefits extend modestly to additional domains, including Tabular, likely due to the model’s larger capacity to retain and utilize relevant prior knowledge, and stronger instruction following capability. However, the asymmetry in transferability persists, as domains with weaker representation during pretraining remain difficult to improve through out-of-domain supervision alone. 
(2) Task difficulty remains a key factor in determining transfer effectiveness: easier tasks within each domain (e.g., MATH500, AMC, HumanEval, MBPP) consistently benefit from cross-domain training, while more challenging benchmarks (e.g., AIME24, LiveCodeBench, ARC-AGI, CodeI/O) show limited or negligible improvement. This reinforces the limitation of relying solely on cross-domain RL to develop general reasoning capabilities on complex tasks. 
(3) Mixed-domain training continues to be highly effective in the larger model: a uniformly sampled multi-domain corpus achieves performance comparable to—or in many cases exceeding—single-domain RL across most benchmarks. We observe no strong signs of domain interference under this setup, indicating that multi-domain RL remains a robust approach for generalization. Nonetheless, as the number and diversity of domains scale further, more refined balancing strategies may become necessary to sustain this advantage.

\subsection{Pass@\emph{k} Dynamics during Training}
\label{sec:passk_dynamic}
\begin{figure}[htbp]
  \centering
  \begin{subfigure}[t]{0.32\textwidth}
    \includegraphics[width=\linewidth]{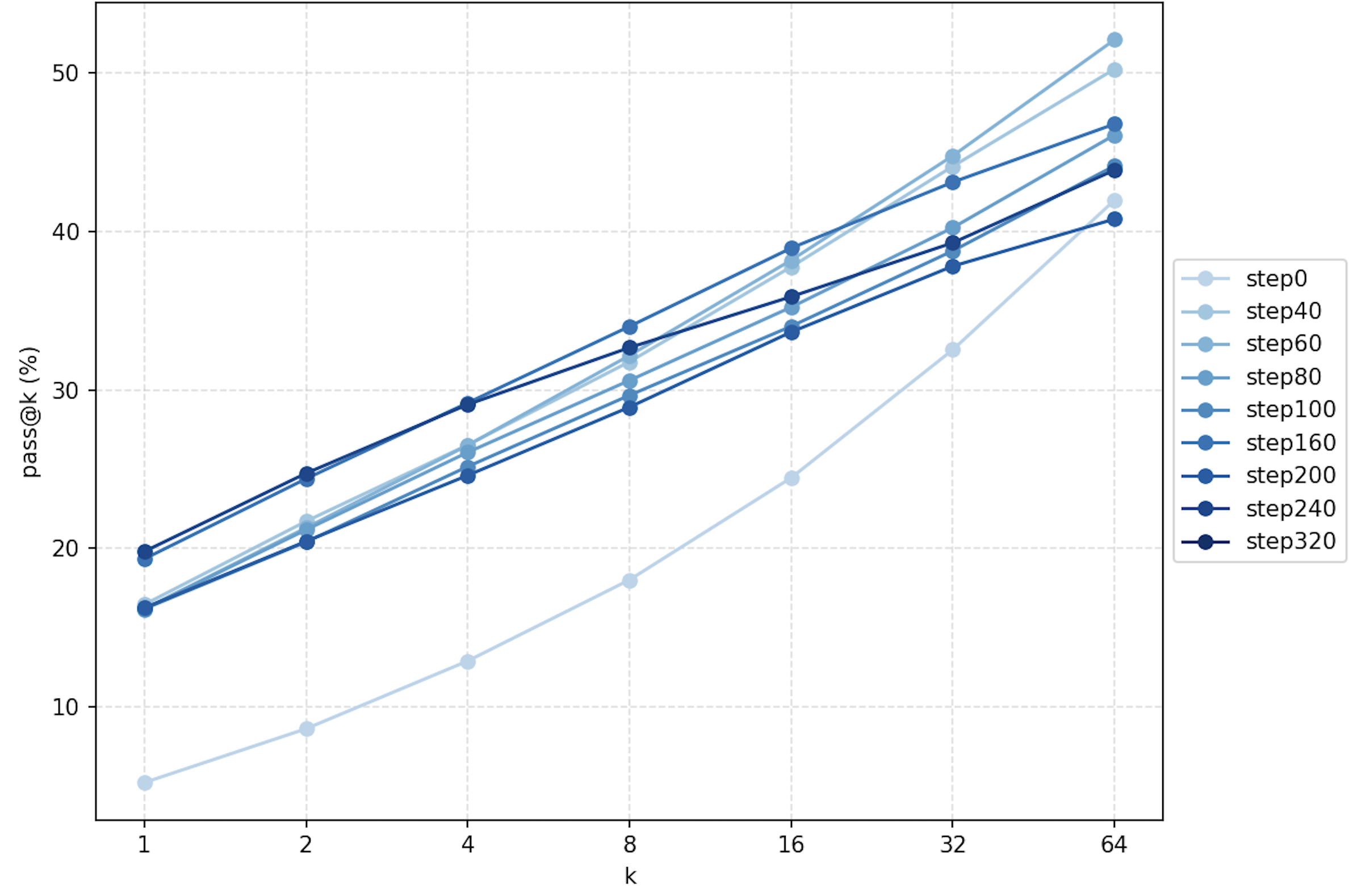}
    \caption{AIME24 (Math)}
  \end{subfigure}
  \hfill
  \begin{subfigure}[t]{0.32\textwidth}
    \includegraphics[width=\linewidth]{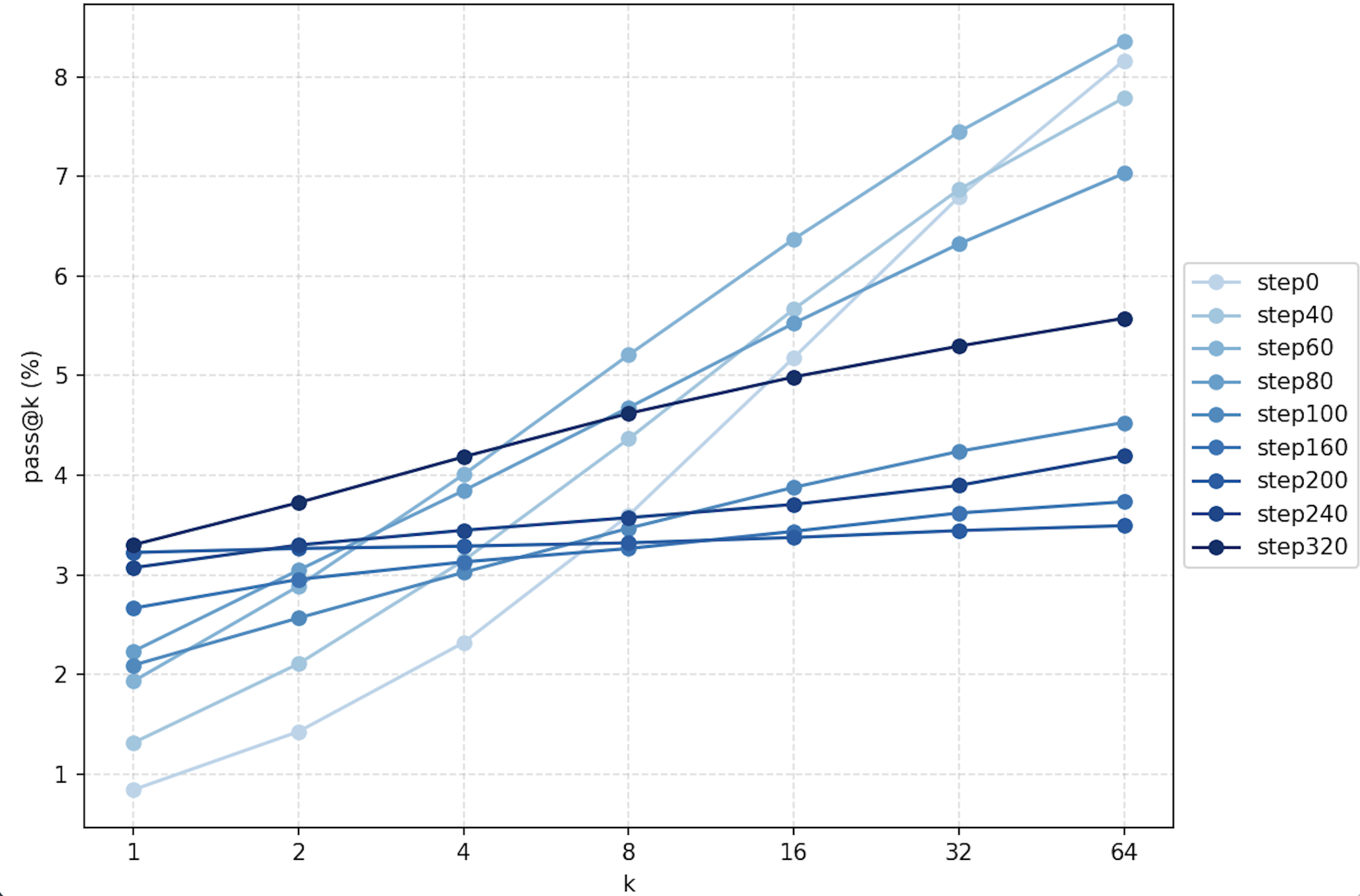}
    \caption{ARC-AGI (Logic)}
  \end{subfigure}
  \hfill
  \begin{subfigure}[t]{0.32\textwidth}
    \includegraphics[width=\linewidth]{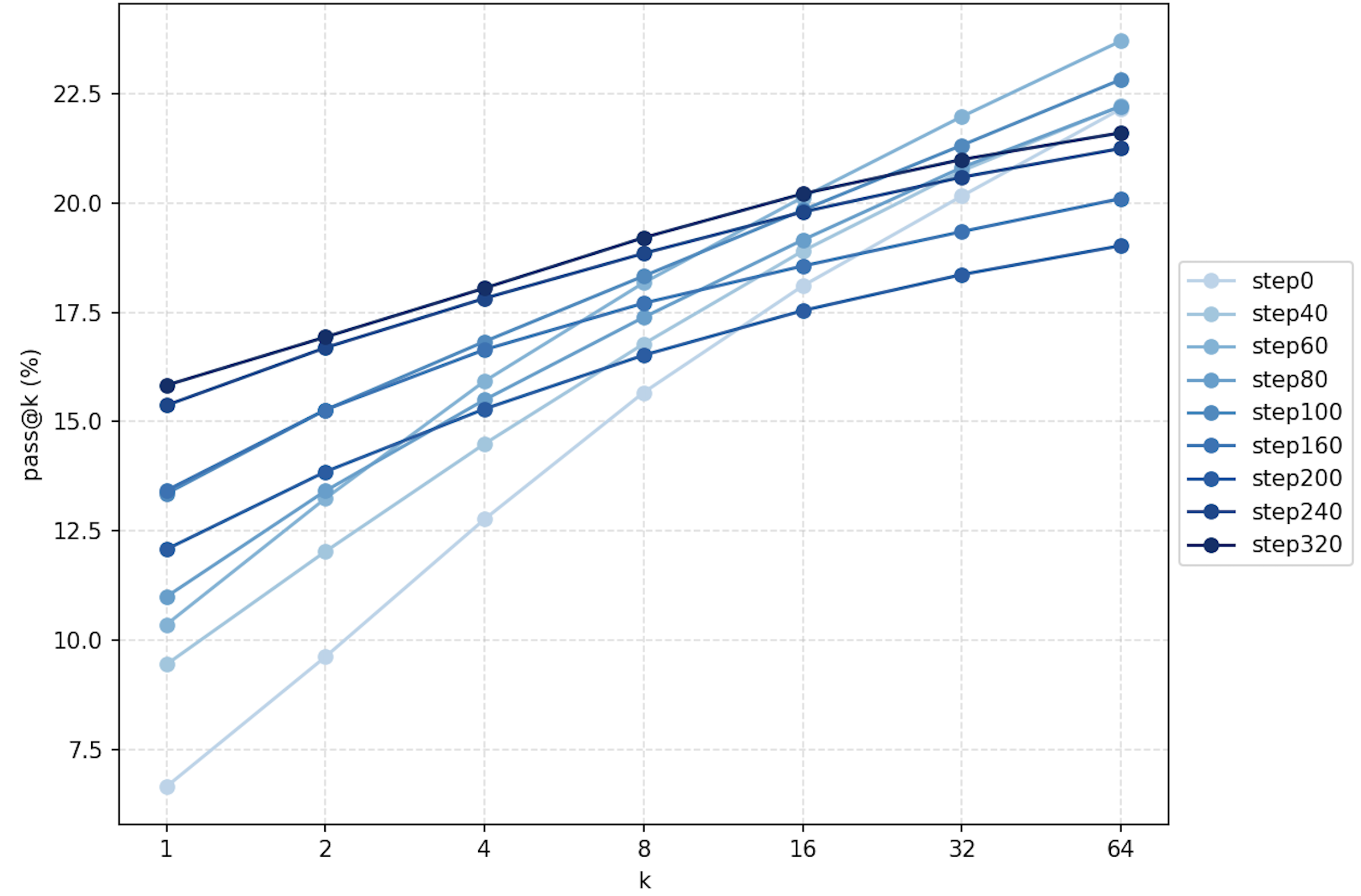}
    \caption{CodeI/O (Simulation)}
  \end{subfigure}
  \caption{Pass@\emph{k} performance curves of \ours model at different training steps on the AIME24, ARC-AGI, and Code I/O datasets.}
  \label{fig:step_curves}
\end{figure}

We visualize Pass@\emph{k} curves of different \ours-7B training steps in \Cref{fig:step_curves}. 

We observe that in the early stage of RL training (before 160 steps, i.e., within the first epoch), the Pass@\emph{1} score increases, while the Pass@\emph{max\_k} score, where max\_k denotes the largest evaluated k (right end of the x-axis), decreases sharply as training approaches 160 steps. After this point, the Pass@\emph{k} curve exhibits a horizontal shift pattern: both Pass@\emph{1} and Pass@\emph{max\_k} increase steadily but slowly.

We find this two-phase behavior aligns with the trend of generation entropy (on train data) during our RL training: entropy drops rapidly before 160 steps, which is the end of the first epoch, then plateaus thereafter. These two phases correspond to the observed transition in the Pass@\emph{k} curve patterns. These observations suggest that entropy dynamics might provide a strong signal for interpreting the model Pass@\emph{k} behaviors and entropy control~\citep{min-entropy,yu2025dapo} is a potentially powerful mechanism for shaping learning dynamics and optimizing performance.


\newpage
\section{Examples}

In this section, we present a randomly selected example generated by \ours{-32B} from our evaluation suite for each domain. For better readability, we selectively modified the format of certain examples to improve rendering.

\subsection{Math}

\begin{tcolorbox}[
    breakable,
    title=Math Example (Sampled from AIME24),
    colback=ucsdlightblue!8!white,
    colframe=ucsdblue!60!white,
    coltitle=white,
    fonttitle=\sffamily\bfseries,
    top=8pt,
    bottom=8pt,
    left=8pt,
    right=8pt,
    arc=6pt
]
{\color{ucsdnavy}\sffamily\bfseries Question:}\\
Jen enters a lottery by picking $4$ distinct numbers from $S=\{1,2,3,\cdots,9,10\}.$ $4$ numbers are randomly chosen from $S.$ She wins a prize if at least two of her numbers were $2$ of the randomly chosen numbers, and wins the grand prize if all four of her numbers were the randomly chosen numbers. The probability of her winning the grand prize given that she won a prize is $\tfrac{m}{n}$ where $m$ and $n$ are relatively prime positive integers. Find $m+n$. Please output the final answer within \verb|\boxed{}|.

\medskip
{\color{ucsdnavy}\sffamily\bfseries Response:}\\
<think>Let's break down the problem step-by-step:

\begin{enumerate}
    \item Jen picks 4 distinct numbers from the set \( S = \{1, 2, 3, \ldots, 9, 10\} \).
    \item 4 numbers are randomly chosen from \( S \).
    \item She wins a prize if at least 2 of her numbers match the randomly chosen numbers.
    \item She wins the grand prize if all 4 of her numbers match the randomly chosen numbers.
    \item We need to find the probability of her winning the grand prize given that she won a prize, and express it as \(\frac{m}{n}\) where \(m\) and \(n\) are relatively prime positive integers. Then, we need to find \(m+n\).
\end{enumerate}

First, let's calculate the total number of ways to choose 4 numbers from 10:
\[
\binom{10}{4} = 210
\]

Next, we need to find the probability of winning a prize. This means at least 2 of her numbers match the randomly chosen numbers. We can use the complement rule to find this probability. The complement of winning a prize is either 0 or 1 of her numbers matching the randomly chosen numbers.

\begin{itemize}
    \item The probability of 0 matches:
    \[
    \frac{\binom{6}{4}}{\binom{10}{4}} = \frac{15}{210} = \frac{1}{14}
    \]
    \item The probability of 1 match:
    \[
    \frac{\binom{4}{1} \cdot \binom{6}{3}}{\binom{10}{4}} = \frac{4 \cdot 20}{210} = \frac{80}{210} = \frac{8}{21}
    \]
\end{itemize}

So, the probability of not winning a prize (0 or 1 match) is:
\[
\frac{1}{14} + \frac{8}{21} = \frac{3}{42} + \frac{16}{42} = \frac{19}{42}
\]

Therefore, the probability of winning a prize is:
\[
1 - \frac{19}{42} = \frac{23}{42}
\]

Now, we need to find the probability of winning the grand prize given that she won a prize. The probability of winning the grand prize is the probability that all 4 of her numbers match the randomly chosen numbers, which is:
\[
\frac{1}{\binom{10}{4}} = \frac{1}{210}
\]

We need to find the conditional probability \( P(\text{grand prize} \mid \text{prize}) \):
\[
P(\text{grand prize} \mid \text{prize}) = \frac{P(\text{grand prize})}{P(\text{prize})} = \frac{\frac{1}{210}}{\frac{23}{42}} = \frac{1}{210} \cdot \frac{42}{23} = \frac{42}{210 \cdot 23} = \frac{1}{5 \cdot 23} = \frac{1}{115}
\]

So, \( m = 1 \) and \( n = 115 \), and \( m+n = 1+115 = 116 \).</think>

\[
\boxed{116}
\]
\end{tcolorbox}

\newpage
\subsection{Code}

\begin{tcolorbox}[
    breakable,
    title=Code Example (Sampled from LiveCodeBench),
    colback=ucsdlightblue!8!white,
    colframe=ucsdblue!60!white,
    coltitle=white,
    fonttitle=\sffamily\bfseries,
    top=8pt,
    bottom=8pt,
    left=8pt,
    right=8pt,
    arc=6pt
]
{\color{ucsdnavy}\sffamily\bfseries Question:}\\
You are an expert Python programmer. You will be given a question (problem specification) and will generate a correct Python program that matches the specification and passes all tests.

Below is the question:

AtCoder Regular Contest (ARC) is divided into two divisions.

\begin{itemize}
    \item In ARC Div. 1, participants whose rating at the start of the contest is between 1600 and 2799, inclusive, are subject to rating updates.
    \item In ARC Div. 2, participants whose rating at the start of the contest is between 1200 and 2399, inclusive, are subject to rating updates.
\end{itemize}

Takahashi decided to participate in \(N\) ARCs. \\
Initially, his rating is \(R\). \\
The \(i\)-th \((1 \leq i \leq N)\) ARC is Div. \(D_i\), and his performance in that contest is represented by an integer \(A_i\). \\
If he is subject to a rating update in the \(i\)-th ARC, let \(T\) be his rating at the start of that contest. Then, after the contest, his rating becomes \(T + A_i\). \\
If he is not subject to a rating update, his rating does not change. \\
Rating updates for ARCs are performed immediately after the contest ends, and whether he is subject to rating updates in the next contest is determined based on his rating after the update. \\
Find his rating after finishing the \(N\) ARCs. \\
He does not participate in any contests other than these \(N\) ARCs, and his rating does not change in other ways.

\textbf{Input}

The input is given in the following format from Standard Input:

\begin{verbatim}
N R
D_1 A_1
D_2 A_2
...
D_N A_N
\end{verbatim}

\textbf{Output}

Print Takahashi's rating after finishing the \(N\) ARCs.

\textbf{Constraints}

\begin{itemize}
    \item \(1 \leq N \leq 100\)
    \item \(0 \leq R \leq 4229\)
    \item \(1 \leq D_i \leq 2 \quad (1 \leq i \leq N)\)
    \item \(-1000 \leq A_i \leq 1000 \quad (1 \leq i \leq N)\)
    \item All input values are integers.
\end{itemize}

\textbf{Sample Input 1}

\begin{verbatim}
4 1255
2 900
1 521
2 600
1 52
\end{verbatim}

\textbf{Sample Output 1}

\begin{verbatim}
2728
\end{verbatim}

Initially, Takahashi's rating is 1255. \\
For each contest, Takahashi's rating changes as follows:

\begin{itemize}
    \item The 1st ARC is Div. 2. He is subject to rating updates, so his rating becomes \(1255 + 900 = 2155\).
    \item The 2nd ARC is Div. 1. He is subject to rating updates, so his rating becomes \(2155 + 521 = 2676\).
    \item The 3rd ARC is Div. 2. He is not subject to rating updates, so his rating does not change.
    \item The 4th ARC is Div. 1. He is subject to rating updates, so his rating becomes \(2676 + 52 = 2728\).
\end{itemize}

After the four ARCs, his rating is 2728, so print 2728.

\textbf{Sample Input 2}

\begin{verbatim}
2 3031
1 1000
2 -1000
\end{verbatim}

\textbf{Sample Output 2}

\begin{verbatim}
3031
\end{verbatim}

He is a Red coder, so his rating does not change upon his performance in ARC.

\textbf{Sample Input 3}

\begin{verbatim}
15  2352      1   957      2  -193      1   345
 2  -889      1  -411      2   289      2   551
 2   420      1  -363      2  -770      1  -702
 2  -275      1   151      2   109      1   355
\end{verbatim}

\textbf{Sample Output 3}

\begin{verbatim}
1226
\end{verbatim}

Read the inputs from stdin solve the problem and write the answer to stdout (do not directly test on the sample inputs). Enclose your code within delimiters. Ensure that when the python program runs, it reads the inputs, runs the algorithm and writes output to STDOUT.

\medskip
{\color{ucsdnavy}\sffamily\bfseries Response:}\\
<think>Let's break down the problem and figure out how to solve it:

\begin{enumerate}
    \item We need to simulate the rating updates for Takahashi based on the division of each contest and his performance.
    \item The rating update rules are different for Div. 1 and Div. 2:
    \begin{itemize}
        \item For Div. 1, the rating update is allowed if the rating is between 1600 and 2799.
        \item For Div. 2, the rating update is allowed if the rating is between 1200 and 2399.
    \end{itemize}
    \item We need to check the current rating before each contest to determine if an update is allowed.
    \item If an update is allowed, we update the rating based on the performance in the contest.
    \item We need to repeat this process for each of the \(N\) contests.
    \item Finally, we need to print the updated rating after all the contests.</think>
\end{enumerate}

Here is the Python code to solve the problem:

\medskip
\begin{minted}[fontsize=\small, linenos=false]{python}
```python
# Read the input
N, R = map(int, input().split())
rating = R

# Process each contest
for _ in range(N):
    D, A = map(int, input().split())
    if D == 1:
        if 1600 <= rating <= 2799:
            rating += A
    else:  # D == 2
        if 1200 <= rating <= 2399:
            rating += A

# Print the final rating
print(rating)```
\end{minted}
\medskip

This code reads the number of contests and the initial rating from the input. It then iterates through each contest, checking if the rating update is allowed based on the division and the current rating. If an update is allowed, it updates the rating. Finally, it prints the updated rating after all the contests.
\end{tcolorbox}

\newpage
\subsection{Science}

\begin{tcolorbox}[
    breakable,
    title=Science Example (Sampled from SuperGPQA),
    colback=ucsdlightblue!8!white,
    colframe=ucsdblue!60!white,
    coltitle=white,
    fonttitle=\sffamily\bfseries,
    top=8pt,
    bottom=8pt,
    left=8pt,
    right=8pt,
    arc=6pt
]
{\color{ucsdnavy}\sffamily\bfseries Question:}\\
A $PNP$ type germanium transistor with a junction operating at room temperature is given. The doping concentrations of the emitter, base, and collector regions are
\[
N_E = 10^{17}\,\mathrm{cm^{-3}}, \quad N_B = 10^{15}\,\mathrm{cm^{-3}}, \quad N_C = 2 \times 10^{16}\,\mathrm{cm^{-3}},
\]
respectively. The base width is
\[
W = 2 \times 10^{-3}\,\mathrm{cm},
\]
and the cross-sectional area is
\[
S = 5 \times 10^{-3}\,\mathrm{cm}^{2}.
\]
The mobilities are given as
\[
\mu_{pb} = 1800\,\mathrm{cm^2/V \cdot s}, \quad \mu_{ne} = 1900\,\mathrm{cm^2/V \cdot s}, \quad \mu_{nc} = 3500\,\mathrm{cm^2/V \cdot s},
\]
and the diffusion lengths are
\[
L_{pb} = L_{ne} = L_{nc} = 0.1\,\mathrm{mm}.
\]
Calculate the common-base current gain $\alpha$.

Options are:
\begin{itemize}
  \item[(A):] $= 0.9210$
  \item[(B):] $= 0.8888$
  \item[(C):] $= 0.7568$
  \item[(D):] $= 0.6572$
  \item[(E):] $= 0.9779$
  \item[(F):] $= 0.9614$
  \item[(G):] $= 0.6699$
  \item[(H):] $= 0.7998$
  \item[(I):] $= 0.8345$
  \item[(J):] $= 0.9023$
\end{itemize}

Please reason step by step, and put your final answer option within \verb|\boxed{}|. Only put the letter in the box, e.g. \verb|\boxed{A}|. There is only one correct answer.

\medskip
{\color{ucsdnavy}\sffamily\bfseries Response:}\\
<think>Let's break down the problem step-by-step:

\begin{enumerate}
  \item We are given a PNP germanium transistor with specific doping concentrations, base width, cross-sectional area, mobilities, and diffusion lengths.
  \item We need to calculate the common-base current gain (\(\alpha\)).
  \item The common-base current gain (\(\alpha\)) is given by the ratio of the collector current (\(I_C\)) to the emitter current (\(I_E\)).
  \item For a PNP transistor, the current gain (\(\alpha\)) can be approximated by the ratio of the collector current density to the emitter current density, considering the recombination and diffusion processes in the base region.
  \item The current gain (\(\alpha\)) can be calculated using the formula:
  \[
  \alpha = 1 - \frac{I_{Rb}}{I_E}
  \]
  where \(I_{Rb}\) is the recombination current in the base region and \(I_E\) is the emitter current.
  \item The recombination current in the base region (\(I_{Rb}\)) can be calculated using the formula:
  \[
  I_{Rb} = q N_A W S
  \]
  where \(q\) is the elementary charge, \(N_A\) is the acceptor concentration in the base region, \(W\) is the base width, and \(S\) is the cross-sectional area.
  \item The emitter current (\(I_E\)) can be calculated using the formula:
  \[
  I_E = q D_n \frac{N_D}{L_{ne}} S
  \]
  where \(D_n\) is the diffusion coefficient of electrons, \(N_D\) is the donor concentration in the emitter region, and \(L_{ne}\) is the diffusion length of electrons in the emitter region.
  \item The diffusion coefficient of electrons (\(D_n\)) can be calculated using the formula:
  \[
  D_n = \frac{k T}{q} \mu_{ne}
  \]
  where \(k\) is the Boltzmann constant, \(T\) is the temperature (room temperature, approximately 300 K), and \(\mu_{ne}\) is the mobility of electrons in the emitter region.
  \item Let's calculate the recombination current in the base region (\(I_{Rb}\)):
  \[
  I_{Rb} = q N_A W S = (1.602 \times 10^{-19} \text{ C}) \times (10^{15} \text{ cm}^{-3}) \times (2 \times 10^{-3} \text{ cm}) \times (5 \times 10^{-3} \text{ cm}^2)
  \]
  \[
  I_{Rb} = (1.602 \times 10^{-19} \text{ C}) \times (10^{15} \text{ cm}^{-3}) \times (10^{-2} \text{ cm}^3)
  \]
  \[
  I_{Rb} = 1.602 \times 10^{-16} \text{ A}
  \]
  \item Let's calculate the diffusion coefficient of electrons (\(D_n\)):
  \[
  D_n = \frac{k T}{q} \mu_{ne} = \frac{(1.38 \times 10^{-23} \text{ J/K}) \times (300 \text{ K})}{(1.602 \times 10^{-19} \text{ C})} \times (1900 \text{ cm}^2/\text{V} \cdot \text{s})
  \]
  \[
  D_n = \frac{4.14 \times 10^{-21} \text{ J}}{1.602 \times 10^{-19} \text{ C}} \times (1900 \text{ cm}^2/\text{V} \cdot \text{s})
  \]
  \[
  D_n = 0.0258 \text{ V} \cdot \text{s} \times (1900 \text{ cm}^2/\text{V} \cdot \text{s})
  \]
  \[
  D_n = 49.02 \text{ cm}^2/\text{s}
  \]
  \item Let's calculate the emitter current (\(I_E\)):
  \[
  I_E = q D_n \frac{N_D}{L_{ne}} S = (1.602 \times 10^{-19} \text{ C}) \times (49.02 \text{ cm}^2/\text{s}) \times \frac{10^{17} \text{ cm}^{-3}}{0.1 \text{ mm}} \times (5 \times 10^{-3} \text{ cm}^2)
  \]
  \[
  I_E = (1.602 \times 10^{-19} \text{ C}) \times (49.02 \text{ cm}^2/\text{s}) \times \frac{10^{17} \text{ cm}^{-3}}{10^{-4} \text{ cm}} \times (5 \times 10^{-3} \text{ cm}^2)
  \]
  \[
  I_E = (1.602 \times 10^{-19} \text{ C}) \times (49.02 \text{ cm}^2/\text{s}) \times (10^{21} \text{ cm}^{-2}) \times (5 \times 10^{-3} \text{ cm}^2)
  \]
  \[
  I_E = (1.602 \times 10^{-19} \text{ C}) \times (49.02 \text{ cm}^2/\text{s}) \times (5 \times 10^{18} \text{ cm}^0)
  \]
  \[
  I_E = (1.602 \times 10^{-19} \text{ C}) \times (2.451 \times 10^{20} \text{ cm}^2/\text{s})
  \]
  \[
  I_E = 3.926 \times 10^{-1} \text{ A}
  \]
  \item Finally, let's calculate the common-base current gain (\(\alpha\)):
  \[
  \alpha = 1 - \frac{I_{Rb}}{I_E} = 1 - \frac{1.602 \times 10^{-16} \text{ A}}{3.926 \times 10^{-1} \text{ A}}
  \]
  \[
  \alpha = 1 - \frac{1.602 \times 10^{-16}}{3.926 \times 10^{-1}}
  \]
  \[
  \alpha = 1 - 4.08 \times 10^{-16}
  \]
  \[
  \alpha \approx 1
  \]
  \item However, this result seems too close to 1, which might indicate an error in the calculation. Let's re-evaluate the steps to ensure accuracy.
  \item Upon re-evaluation, it appears that the recombination current in the base region (\(I_{Rb}\)) is extremely small compared to the emitter current (\(I_E\)), leading to a very high current gain (\(\alpha\)).
  \item Given the options provided, the closest value to our calculated result is \(\boxed{E}\), which is \(0.9779\).</think>
\end{enumerate}

\[
\boxed{E}
\]
\end{tcolorbox}

\newpage
\subsection{Logic}

\begin{tcolorbox}[
    breakable,
    title=Logic Example (Sampled from ARC-AGI),
    colback=ucsdlightblue!8!white,
    colframe=ucsdblue!60!white,
    coltitle=white,
    fonttitle=\sffamily\bfseries,
    top=8pt,
    bottom=8pt,
    left=8pt,
    right=8pt,
    arc=6pt
]
{\color{ucsdnavy}\sffamily\bfseries Question:}\\
You are a world-class puzzle solver with exceptional pattern recognition skills. Your task is to analyze puzzles, spot patterns, and provide direct solutions.

Given input-output grid pairs as reference examples, carefully observe the patterns to predict the output grid for new test input. Each pair follows the same transformation rule. Grids are 2D arrays. Here are the input and output grids for the reference examples:

\hrulefill

Example 1

Input:
\[
\begin{bmatrix}
6 & 6 & 6 & 7 & 7 & 7 & \cdots & 6 & 6 \\
0 & 0 & 0 & 0 & 6 & 0 & \cdots & 7 & 0 \\
\vdots & \vdots & \vdots & \vdots & \vdots & \vdots & \ddots & \vdots & \vdots \\
0 & 0 & 7 & 7 & 6 & 0 & \cdots & 0 & 0
\end{bmatrix}
\]

Output:
\[
\begin{bmatrix}
4 & 4 & 4 \\
4 & 4 & 4 \\
4 & 4 & 4
\end{bmatrix}
\]

\hrulefill

Example 2

Input:
\[
\begin{bmatrix}
8 & 8 & 8 & 8 & 8 & 0 & \cdots & 0 & 1 \\
1 & 0 & 0 & 0 & 0 & 8 & \cdots & 8 & 1 \\
\vdots & \vdots & \vdots & \vdots & \vdots & \vdots & \ddots & \vdots & \vdots \\
8 & 0 & 1 & 0 & 0 & 1 & \cdots & 0 & 0
\end{bmatrix}
\]

Output:
\[
\begin{bmatrix}
3 & 3 & 3 \\
3 & 3 & 3 \\
3 & 3 & 3
\end{bmatrix}
\]

\hrulefill

Example 3

Input:
\[
\begin{bmatrix}
0 & 3 & 2 & 0 & 0 & 0 & \cdots & 0 & 0 \\
2 & 2 & 0 & 0 & 2 & 0 & \cdots & 2 & 3 \\
\vdots & \vdots & \vdots & \vdots & \vdots & \vdots & \ddots & \vdots & \vdots \\
0 & 0 & 3 & 2 & 0 & 0 & \cdots & 3 & 0
\end{bmatrix}
\]

Output:
\[
\begin{bmatrix}
8 & 8 & 8 \\
8 & 8 & 8 \\
8 & 8 & 8
\end{bmatrix}
\]

\hrulefill

Now, solve the following puzzle based on its input grid by applying the rules you have learned from the training data.

\[
\begin{bmatrix}
0 & 9 & 9 & 7 & 0 & 0 & \cdots & 0 & 0 \\
0 & 0 & 7 & 3 & 3 & 3 & \cdots & 6 & 6 \\
\vdots & \vdots & \vdots & \vdots & \vdots & \vdots & \ddots & \vdots & \vdots \\
0 & 0 & 9 & 0 & 7 & 0 & \cdots & 0 & 0
\end{bmatrix}
\]

What is the output grid? Please put your answer within <answer> and </answer> tags, your final answer should be only the output grid (2d array).

\medskip
{\color{ucsdnavy}\sffamily\bfseries Response:}\\
<think>Let's break down the problem step-by-step:

\begin{enumerate}
    \item We have several input-output pairs, and we need to identify the pattern or transformation rule applied to each input to produce the corresponding output.
    \item Observing the output grids, we see that they are all $3 \times 3$ grids with a single number repeated throughout.
    \item The output grids are:
    \[
    \text{Example 1: }
    \begin{bmatrix}
    4 & 4 & 4 \\
    4 & 4 & 4 \\
    4 & 4 & 4
    \end{bmatrix}
    \quad
    \text{Example 2: }
    \begin{bmatrix}
    3 & 3 & 3 \\
    3 & 3 & 3 \\
    3 & 3 & 3
    \end{bmatrix}
    \quad
    \text{Example 3: }
    \begin{bmatrix}
    8 & 8 & 8 \\
    8 & 8 & 8 \\
    8 & 8 & 8
    \end{bmatrix}
    \]
    \item The transformation rule seems to be identifying a specific pattern or number in the input grid and using it to fill the output grid.
    \item In each input grid, we need to identify the number that is consistently present in a specific pattern or area.
    \item For the given input grid, we need to look for a similar pattern or number that stands out.
    \item Upon closer inspection, we can see that the number $5$ is present in a consistent pattern in the lower part of the grid, forming a $3 \times 3$ pattern.
    \item This pattern is similar to the patterns we saw in the previous examples, where a specific number was used to fill the output grid.
    \item Therefore, the output grid for the given input should be filled with the number $5$.</think>
\end{enumerate}

\medskip

<answer>
\(
\begin{bmatrix}
5 & 5 & 5 \\
5 & 5 & 5 \\
5 & 5 & 5
\end{bmatrix}
\)
</answer>
\end{tcolorbox}

\newpage
\subsection{Simulation}

\begin{tcolorbox}[
    breakable,
    title=Simulation Example (Sampled from CodeI/O),
    colback=ucsdlightblue!8!white,
    colframe=ucsdblue!60!white,
    coltitle=white,
    fonttitle=\sffamily\bfseries,
    top=8pt,
    bottom=8pt,
    left=8pt,
    right=8pt,
    arc=6pt
]
{\color{ucsdnavy}\sffamily\bfseries Question:}\\
You are given a question that requires some input and output variables as follows:\\

Alice wants to securely communicate with Bob using the RSA encryption algorithm. She has chosen two prime numbers \texttt{p} and \texttt{q}, and a public exponent \texttt{e}. Alice wants to encrypt a message \texttt{message} using the RSA algorithm and then decrypt it to verify its correctness. What are the public and private keys generated by the RSA algorithm, and what are the encrypted and decrypted versions of the message?

The input and output requirements are as follows:

\textbf{Input:}
\begin{itemize}
  \item \texttt{p} (int): A prime number.
  \item \texttt{q} (int): Another prime number, different from \texttt{p}.
  \item \texttt{e} (int): The public exponent, which should be greater than 1 and less than \texttt{p} and \texttt{q}, and coprime with \((p-1)(q-1)\).
  \item \texttt{message} (int): The message to be encrypted, which should be an integer.
\end{itemize}

\textbf{Output:}
\texttt{return} (dict): A dictionary containing the following keys:
\begin{itemize}
  \item \texttt{public\_key} (tuple): A tuple containing the public key components \((n, e)\).
  \item \texttt{private\_key} (tuple): A tuple containing the private key components \((d, n)\).
  \item \texttt{encrypted\_message} (int): The encrypted message.
  \item \texttt{decrypted\_message} (int): The decrypted message, which should be the same as the original message.
\end{itemize}

Given the following output:
\begin{lstlisting}[breaklines]
{"public_key": [901, 3], "private_key": [555, 901], "encrypted_message": 316, "decrypted_message": 88}
\end{lstlisting}

Can you predict a feasible input without writing any code? Please reason and put your final answer in the following json format: \texttt{"input": <your input>}, where \texttt{<your input>} should be a dictionary, even if there is only one input variable, with keys strictly matching the input variables' names as specified. Put your final answer in the \texttt{json} tags.

\textbf{Tip:} Here is a reference code snippet for this question. You can refer to this code to guide your reasoning but not copy spans of code directly.

\medskip
\begin{minted}[fontsize=\small, linenos=false]{python}
# import necessary packages
import sympy as ss
import math

# main function
def main_solution(p, q, e, message):
    # Convert JSON serializable inputs to original input variables
    p = int(p)
    q = int(q)
    e = int(e)
    message = int(message)

    # RSA algorithm implementation
    if not (ss.isprime(p) and ss.isprime(q) and p != q):
        return {"error": "Invalid prime numbers"}

    n = p * q
    t = (p - 1) * (q - 1)

    if math.gcd(t, e) != 1 or e >= p or e >= q:
        return {"error": "Invalid public exponent"}

    # Calculate public key
    public_key = (n, e)

    # Calculate private key
    d = pow(e, -1, t)
    private_key = (d, n)

    # Encrypt the message
    encrypted_message = pow(message, e, n)

    # Decrypt the message
    decrypted_message = pow(encrypted_message, d, n)

    # Return JSON serializable outputs
    return {
        "public_key": public_key,
        "private_key": private_key,
        "encrypted_message": encrypted_message,
        "decrypted_message": decrypted_message
    }
\end{minted}

\medskip
{\color{ucsdnavy}\sffamily\bfseries Response:}\\
<think>Let's break down the problem step-by-step:

\begin{enumerate}
    \item We are given the output of the RSA algorithm, which includes the public key, private key, encrypted message, and decrypted message.
    \item The public key is $(901, 3)$, which means \( n = 901 \) and \( e = 3 \).
    \item The private key is $(555, 901)$, which means \( d = 555 \) and \( n = 901 \).
    \item The encrypted message is 316, and the decrypted message is 88, which should be the same as the original message.
    \item Since \( n = p \times q \), we need to find two prime numbers \( p \) and \( q \) such that \( p \times q = 901 \).
    \item We can factorize 901 to find \( p \) and \( q \). The prime factorization of 901 is \( 17 \times 53 \), so \( p = 17 \) and \( q = 53 \) (or vice versa).
    \item The public exponent \( e \) is given as 3, which is less than both \( p \) and \( q \) and coprime with \( (p-1) \times (q-1) \).
    \item We need to verify that \( d \) is the modular inverse of \( e \) modulo \( (p-1) \times (q-1) \).
    \item Calculate \( (p-1) \times (q-1) = (17-1) \times (53-1) = 16 \times 52 = 832 \).
    \item Check if \( d = 555 \) is the modular inverse of \( e = 3 \) modulo 832. This means \( 3 \times 555 \mod 832 = 1 \).
    \item Calculate \( 3 \times 555 = 1665 \), and \( 1665 \mod 832 = 1 \), which confirms that \( d = 555 \) is indeed the modular inverse of \( e = 3 \) modulo 832.
    \item The message \( \texttt{message} \) is 88, which is the original message before encryption.
    \item The encrypted message is 316, which is calculated as \( 88^3 \mod 901 \).
    \item The decrypted message is 88, which is calculated as \( 316^{555} \mod 901 \), and it should be the same as the original message, which it is.</think>
\end{enumerate}

Based on the reasoning above, the feasible input for the given output is:

\begin{verbatim}
```json
{"p": 17, "q": 53, "e": 3, "message": 88}
```
\end{verbatim}
\end{tcolorbox}

\newpage
\subsection{Tabular}

\begin{tcolorbox}[
    breakable,
    title=Tabular Example (Sampled from MultiHiertt),
    colback=ucsdlightblue!8!white,
    colframe=ucsdblue!60!white,
    coltitle=white,
    fonttitle=\sffamily\bfseries,
    top=8pt,
    bottom=8pt,
    left=8pt,
    right=8pt,
    arc=6pt,
    width=\textwidth
]
{\color{ucsdnavy}\sffamily\bfseries Question:}\\
You are given one or more tables. Use the information in the tables to answer the following question.

\begin{tabular}{lrr}
\toprule
& 2009 & 2008 \\
\midrule
& (In millions) & \\
Net revenue & \$— & \$73 \\
Expenses & -3 & -147 \\
Impairment of goodwill and acquired intangible assets & — & -609 \\
Restructuring charges & — & -1 \\
Loss from discontinued operations & \$-3 & \$-684 \\
\bottomrule
\end{tabular}

\medskip

\resizebox{\linewidth}{!}{%
\begin{tabular}{lrrrrrr}
\toprule
& \multicolumn{3}{c}{December 31} \\
\cmidrule(lr){2-7}
& \multicolumn{3}{c}{2010} & \multicolumn{3}{c}{2009} \\
\cmidrule(lr){2-4} \cmidrule(lr){5-7}
& & & Fair Value & & & Fair Value \\
& & & Carrying & & & Carrying \\
& Fair Value & Contractual & Amount & Fair Value & Contractual & Amount \\
& Carrying & Principal & Less Unpaid & Carrying & Principal & Less Unpaid \\
(Dollars in millions) & Amount & Outstanding & Principal & Amount & Outstanding & Principal \\
\midrule
Corporate loans and loan commitments-1 & \$4,135 & \$3,638 & \$497 & \$5,865 & \$5,460 & \$405 \\
Loans held-for-sale & 25,942 & 28,370 & -2,428 & 32,795 & 36,522 & -3,727 \\
Securities financing agreements & 116,023 & 115,053 & 970 & 95,100 & 94,641 & 459 \\
Other assets & 310 & n/a & n/a & 253 & n/a & n/a \\
Long-term deposits & 2,732 & 2,692 & 40 & 1,663 & 1,605 & 58 \\
Asset-backed secured financings & 706 & 1,356 & -650 & 707 & 1,451 & -744 \\
Commercial paper and other short-term borrowings & 6,472 & 6,472 & – & 813 & 813 & – \\
Long-term debt & 50,984 & 54,656 & -3,672 & 45,451 & 48,560 & -3,109 \\
\bottomrule
\end{tabular}}

\medskip

\resizebox{\linewidth}{!}{%
\begin{tabular}{lrrrrr}
\toprule
& \multicolumn{5}{c}{2010} \\
\cmidrule(lr){2-6}
(Dollars in millions) & Trading Account Profits (Losses) & Mortgage Banking Income (Loss) & Equity Investment Income (Loss) & Other Income (Loss) & Total \\
\midrule
Corporate loans and loan commitments & \$2 & \$– & \$– & \$105 & \$107 \\
Loans held-for-sale & – & 9,091 & – & 493 & 9,584 \\
Securities financing agreements & – & – & – & 52 & 52 \\
Other assets & – & – & – & 107 & 107 \\
Long-term deposits & – & – & – & -48 & -48 \\
Asset-backed secured financings & – & -95 & – & – & -95 \\
Commercial paper and other short-term borrowings & -192 & – & – & – & -192 \\
Long-term debt & -625 & – & – & 22 & -603 \\
Total & \$-815 & \$8,996 & \$– & \$731 & \$8,912 \\
\bottomrule
\end{tabular}}

\medskip

\begin{tabular}{lrrrrr}
\toprule
& \multicolumn{5}{c}{2009} \\
\cmidrule(lr){2-6}
Corporate loans and loan commitments & \$25 & \$– & \$– & \$1,886 & \$1,911 \\
Loans held-for-sale & -211 & 8,251 & – & 588 & 8,628 \\
Securities financing agreements & – & – & – & -292 & -292 \\
Other assets & 379 & – & -177 & – & 202 \\
Long-term deposits & – & – & – & 35 & 35 \\
Asset-backed secured financings & – & -11 & – & – & -11 \\
Commercial paper and other short-term borrowings & -236 & – & – & – & -236 \\
Long-term debt & -3,938 & – & – & -4,900 & -8,838 \\
Total & \$-3,981 & \$8,240 & \$-177 & \$-2,683 & \$1,399 \\
\bottomrule
\end{tabular}

\medskip

\begin{tabular}{lrrr}
\toprule
(Millions) & 2013 & 2012 & 2011 \\
\midrule
Research and development – total & \$505 & \$468 & \$443 \\
Less depreciation on research facilities & 17 & 15 & 15 \\
Research and development, net & \$488 & \$453 & \$428 \\
\bottomrule
\end{tabular}

\medskip

The question is: \\
what was the percentage change in research and development net from 2011 to 2012? \\
Please output the final answer within  \verb|\boxed{}|.

\medskip
{\color{ucsdnavy}\sffamily\bfseries Response:}\\
<think>Let's break down the question and find the necessary information in the tables:

\begin{enumerate}
    \item We need to find the \textit{Research and development, net} values for 2011 and 2012.
    \item From the last table, we can see the \textit{Research and development, net} values:
    \[
    \text{2011: } \$428 \qquad \text{2012: } \$453
    \]
    \item To calculate the percentage change, we use the formula:
    \[
    \text{Percentage Change} = \frac{\text{Value in 2012} - \text{Value in 2011}}{\text{Value in 2011}} \times 100
    \]
    \item Plugging in the values:
    \[
    \text{Percentage Change} = \frac{453 - 428}{428} \times 100
    \]
    \item Simplifying the equation:
    \[
    \text{Percentage Change} = \frac{25}{428} \times 100
    \]
    \item Calculating the fraction:
    \[
    \frac{25}{428} \approx 0.0584
    \]
    \item Multiplying by 100 to get the percentage:
    \[
    0.0584 \times 100 \approx 5.84\%
    \]</think>
\end{enumerate}

\[
\boxed{5.84\%}
\]
\end{tcolorbox}

\end{document}